\documentclass{article}

\usepackage[preprint]{neurips_2026}

\usepackage[utf8]{inputenc} % allow utf-8 input
\usepackage[T1]{fontenc}    % use 8-bit T1 fonts
\usepackage{hyperref}       % hyperlinks
\usepackage{url}            % simple URL typesetting
\usepackage{booktabs}       % professional-quality tables
\usepackage{amsfonts}       % blackboard math symbols
\usepackage{amssymb}        % \varnothing, \mathbb, etc.
\usepackage{nicefrac}       % compact symbols for 1/2, etc.
\usepackage{microtype}      % microtypography
\usepackage{xcolor}         % colors
\definecolor{cblue}{RGB}{0,90,180}                                                                                                      
\definecolor{enhancementblue}{RGB}{30,144,255} \usepackage{amsmath}
\usepackage{amsthm}
\usepackage{graphicx}
\usepackage{enumitem}
\usepackage{algorithm}
\usepackage{algpseudocode}
\usepackage{tikz}
\usetikzlibrary{shapes,arrows,positioning,fit,backgrounds,decorations.pathreplacing,calc}
\usepackage{wrapfig}

\newtheorem{theorem}{Theorem}
\newtheorem{lemma}{Lemma}

\theoremstyle{definition}

\title{Adaptive Bayesian Partner Selection for Federated Clinical Centers}
\vspace{-10mm}

\author{%
  Navid Seidi \\
  Department of Computer Science\\
  Missouri University of Science and Technology\\
  Rolla, MO, USA \\
  \texttt{nseidi@mst.edu} \\
  \And
  Satyaki Roy \\
  Department of Mathematical Sciences\\
  University of Alabama in Huntsville\\
  Huntsville, AL, USA \\
  \texttt{sr0215@uah.edu} \\
  \And
  Sajal K. Das \\
  Department of Computer Science\\
  Missouri University of Science and Technology\\
  Rolla, MO, USA \\
  \texttt{sdas@mst.edu} \\
}

\begin{document}

\maketitle
\vspace{-8mm}
\begin{abstract}
\vspace{-2mm}

% Version II
Federated learning (FL) in healthcare is challenged by pronounced heterogeneity and temporal concept drift across clinical centers, where evolving patient populations and care practices shift data distributions. Existing approaches rely on persistent global communication, incurring substantial bandwidth overhead while risking negative transfer from poorly aligned peers. We address this by proposing \textsc{Adaptive Bayesian Partner Selection} (\textsc{ABPS}), a peer-to-peer framework that governs \emph{who} collaborates, \emph{when}, and at \emph{what cost}. Each center maintains a Beta–Bernoulli posterior over prospective peers’ Shapley marginal utility, ranks candidates using an Upper Confidence Bound (UCB) criterion, and forms collaborations through a lightweight propose–reject mechanism, with the option to abstain from communication when mutually beneficial interaction fails. It admits a stochastic decision interpretation, yielding finite-sample concentration guarantees and {\small $\mathcal{O}(\kappa \log T)$} regret in partner selection, along with conditions under which intentional isolation is optimal under negative transfer. Lightweight extensions, head personalization, bfloat16 quantized communication, and tunable active-set cardinality further improve efficiency, while a goal-aware metadata filter enables institution-specific collaboration strategies. Evaluations on binary in-hospital mortality prediction over the first 24 hours of an ICU stay, with 230 non-IID clinical centers drawn from MIMIC-IV, show that the full \textsc{ABPS-X} variant matches the strongest federated baseline (FedDyn, AUROC 0.758) at 0.09× the communication cost of FedAvg, with reduced variability. A diversity-driven configuration activates intentional isolation for a substantial fraction of centers, highlighting the role of selective collaboration under heterogeneity. These results show that adaptive, utility-aware collaboration reduces communication without sacrificing accuracy when centers are numerous and small, providing a scalable paradigm for real-world healthcare FL systems.

\end{abstract}
\vspace{-6mm}
\section{Introduction}\label{sec:intro}
\vspace{-4mm}

Federated Learning (FL) has emerged as the dominant paradigm for training predictive models across decentralized centers without exposing sensitive patient data~\cite{mcmahan2017communication}, sidestepping Health Insurance Portability and Accountability Act (HIPAA)-style data-sharing constraints. Realizing FL in clinical settings is nevertheless gated by two persistent obstacles: statistical heterogeneity (i.e., non-IID biomedical data) across institutions~\cite{hsu2019measuring}, and concept drift as patient demographics and clinical protocols evolve~\cite{gama2014survey, rahimli2024federated} (see Appendix~\ref{app:drift_evidence} for an instance of distributional drift across admission eras in routinely charted vital signs).

\vspace{-2mm}
This heterogeneity is structural rather than incidental. The healthcare ecosystem comprises three institution types with divergent key performance indicators (KPIs) and data profiles. \emph{Integrated delivery networks} (Kaiser Permanente, HCA Healthcare) aggregate large but internally siloed cohorts. \emph{Academic medical centers} see complex, rare, and trial-driven cases at low volume. \emph{Community hospitals} generate the majority of high-volume routine encounters. No single ecosystem captures the full distribution required for generalizable modeling, yet forcing central FL aggregation across them frequently induces negative transfer that degrades local performance. Peer-to-peer (P2P) alternatives~\cite{roy2019braintorrent, hegedus2021gossip} avoid the single point of failure but still face the question of \emph{which} peers a center should collaborate with and at what bandwidth cost.\\
% {\color{brown}
% \textbf{Contributions.} To achieve predictive accuracy under concept drift, we introduce \textsc{Adaptive Bayesian Partner Selection} (\textsc{ABPS}), a serverless P2P federated learning framework in which each center adaptively selects collaborators based on a Bayesian estimate of their utility. \textsc{ABPS} models each candidate peer’s Shapley marginal contribution via a Beta--Bernoulli posterior, ranks peers using an $\epsilon$-greedy UCB policy, and employs a propose--reject protocol with an explicit \emph{rest} action to mitigate negative transfer. We establish three guarantees: posterior concentration (Theorem~\ref{thm:convergence}), sublinear regret of order $\mathcal{O}(\kappa \log T)$ (Theorem~\ref{thm:regret}), and the Bayes-optimality of isolation (Lemma~\ref{lem:isolation}). The core framework composes head personalization, bfloat16~\cite{kalamkar2019bfloat} quantized communication, and a tunable active-set cardinality $\kappa$. In addition, a goal-aware metadata pre-filter allows each center to prioritize homogeneity, diversity, or KPI alignment objectives. On MIMIC-IV v3.1, with $n{=}230$ non-IID careunit-by-year centers, \textsc{ABPS-X} matches the strongest federated baseline (FedDyn) while using only $0.09\times$ the bandwidth of FedAvg. In contrast, applying the same extensions to FedAvg degrades performance by $6.4$ AUROC points, leaving \textsc{ABPS-X} ahead of FedAvg-X by $6.7$ points (Table~\ref{tab:headline}), thereby isolating the gains to Bayesian partner-selection.
% }
\textbf{Contributions.} To achieve predictive accuracy under concept drift, we introduce \textsc{Adaptive Bayesian Partner Selection} (\textsc{ABPS}), a serverless P2P federated learning framework in which each center adaptively selects collaborators based on a Bayesian estimate of their utility. \textsc{ABPS} models each candidate peer’s Shapley marginal contribution via a Beta--Bernoulli posterior, ranks peers using an {\small $\epsilon$}-greedy UCB policy, and employs a propose--reject protocol with an explicit \emph{rest} action to mitigate negative transfer. We establish three guarantees: posterior concentration (Theorem~\ref{thm:convergence}), sublinear regret of order {\small $\mathcal{O}(\kappa \log T)$} (Theorem~\ref{thm:regret}), and the Bayes-optimality of isolation (Lemma~\ref{lem:isolation}). The core framework comprises head personalization, bfloat16~\cite{kalamkar2019bfloat} quantized communication, and a tunable active-set cardinality {\small $\kappa$}. In addition, a goal-aware metadata pre-filter allows each center to prioritize homogeneity, diversity, or KPI alignment objectives. We denote the maximally-extended configuration as \textsc{ABPS-X}, which composes all of the above with deeper {\small $2$}-layer head personalization, server-side exponential moving average (EMA) momentum, and validation-AUROC early stopping (full hyperparameters in Sec.~\ref{sec:headline}). On MIMIC-IV v3.1, with {\small $n{=}230$} non-IID careunit-by-year centers, \textsc{ABPS-X} matches the strongest federated baseline (FedDyn) while using only {\small $0.09\times$} the bandwidth of FedAvg. In contrast, applying the same extensions to FedAvg degrades performance by {\small $6.4$} AUROC points, leaving \textsc{ABPS-X} ahead of FedAvg-X by {\small $6.7$} points (Table~\ref{tab:headline}), thereby isolating the gains to Bayesian partner-selection.

\vspace{-4mm}
\section{Related Work}\label{sec:related_work}
\vspace{-4mm}

Our research intersects with several active domains in decentralized machine learning. We organize the related work below in the order in which the corresponding open \textbf{\underline{Q}}uestion is answered in the body of the paper, ranging from divergence-aware partner selection to peer-to-peer (P2P) federated architectures, concept-drift adaptation, and communication-efficient FL.

\vspace{-4mm}
\subsection{Divergence Metrics for Non-IID Collaboration}
\vspace{-3mm}

Information-theoretic divergences (Kullback-Leibler, hereafter KL, and Wasserstein) quantify inter-client distributional disparity in FL~\cite{nemeth2025feddiverse}. They appear as regularization penalties that align local models without raw data exposure~\cite{nemeth2025feddiverse}, as client-selection utilities that favor either stabilizing or diversity-injecting clients~\cite{dusing2025distribution, rahad2025kl-feddis}, and as theoretical tools for bounding FL generalization under heterogeneity. This philosophy extends to the wire-format level: each center can be summarized by a non-PHI descriptor of its distribution (e.g., positive-class rate, training-set size, optional Charlson-comorbidity prevalence), and peers can be admitted via a goal-dependent monotone function of pairwise similarity. A key limitation is that divergence is treated as a single scalar objective fixed network-wide, without allowing individual centers to declare their own collaboration goal before model exchange. Yet healthcare institution types (Integrated Delivery Networks, Academic Medical Centers, Community Hospitals) require different collaboration objectives, including homogeneity, diversity, or KPI alignment~\cite{li2025fedhealthcare}. We label this open problem \textbf{Q1} \textit{(goal-heterogeneous collaboration)} and address it through the goal-aware metadata pre-filter of Sec.~\ref{sec:goal_filter}, where a center selects one of three admission rules, namely, cosine, anti-cosine, and KPI matching, at runtime without risking exposing model parameters.

% The family is operationally restricted but the framework admits richer monotone goals.

\vspace{-4mm}
\subsection{P2P Federated Learning and Partner Selection}
\vspace{-3mm}

Peer-to-peer (P2P) FL decentralizes aggregation, eliminating single points of failure and reducing communication bottlenecks~\cite{hegedus2021gossip, zhou2024defta}. However, identifying the appropriate collaboration topology under non-IID heterogeneity remains an open challenge. Early approaches relied on gossip-based protocols~\cite{hegedus2021gossip}, while more recent work studies the security of P2P training against backdoor attacks~\cite{syros2023backdoor} and robustness mechanisms against free-riding and collusion~\cite{augello2024tackling, ranjan2022securing}. Within the multi-armed bandit (MAB) framework, CS-UCB~\cite{xia2020csucb} introduced UCB-style scheduling in server-mediated FL, and subsequent work extended this idea to decentralized settings with non-stationary MAB formulations over peer groups~\cite{listozec2024efficient}. Complementary approaches based on Shapley value estimate client contributions for server-side selection~\cite{singhal2024greedyshapley, yang2024maverick}. A second limitation persists across P2P, MAB-, and Shapley-based methods: all assume mandatory participation in each round. Even decentralized variants enforce collaboration once the topology is established, despite evidence that aggregation across heterogeneous institutions can degrade local performance below the no-collaboration baseline~\cite{crowson2022recent, li2025fedhealthcare}. We define this as \textbf{Q2} \textit{(selective collaboration under heterogeneity)} and address it through the Bayesian propose–reject mechanism (discussed in Sec.~\ref{sec:topology_subsec}), allowing each center to abstain from participation when all posterior UCB scores fall below the acceptance threshold {\small $\tau_{\mathrm{acc}}$}.

\vspace{-4mm}
\subsection{Concept Drift and Bayesian Adaptation}
\vspace{-3mm}

Clinical data distributions evolve due to changes in patient demographics, treatment protocols, and clinical infrastructure, leading to concept drift~\cite{rahimli2024federated}. Bayesian methods quantify the resulting uncertainty, with two dominant approaches in prior work. The first models uncertainty over \emph{model parameters}, where Bayesian neural networks in FL aggregate posterior distributions rather than point estimates~\cite{saile2024client-side, rahman2025detect}, often incorporating hierarchical or personalized updates to adapt to local data. The second leverages uncertainty for \emph{update filtering}, where client contributions are selectively incorporated based on confidence measures, such as credible-interval thresholds~\cite{iglesias2024twostudents}. Despite their effectiveness, both approaches share a key limitation: abstaining from collaboration is not treated as a principled decision with formal guarantees. In practice, concept drift can render collaboration rounds detrimental, yet existing methods rarely allow nodes to opt out in a theoretically grounded manner~\cite{crowson2022recent, gama2014survey, rahimli2024federated}. We define this as \textbf{Q3} \textit{(adaptive isolation under drift)} and address it through an explicit rest action in Sec.~\ref{sec:topology_subsec}, supported by a Bayes-optimal isolation criterion (Lemma~\ref{lem:isolation}) that provides a closed-form condition under which abstention strictly outperforms collaboration.

\vspace{-4mm}
\subsection{Personalization, Quantization, and Communication-Efficient FL}
\vspace{-3mm}

To achieve personalization and bandwidth minimization, it is necessary to keep a subset of model parameters local per client: FedPer and FedRep retain the classifier head locally~\cite{arivazhagan2019federated}. pFedHN and Ditto provide other per-client adaptations. Quantized transmission compresses the wire format: LLM.int8~\cite{dettmers2022llmint8}, QSGD, and signSGD demonstrate single-digit-percent accuracy loss with {\small $2$-$16\times$} bandwidth reduction. Server momentum (FedAvgM) stabilizes aggregation under heterogeneity. These ideas are orthogonal to partner selection, and in principle, they can be composed with any partner-selection core to push the accuracy-bandwidth frontier further. A fourth limitation runs across all three of these families: bandwidth-saving mechanisms have been studied in isolation from \emph{which} peers a center should engage with. Real-world multi-institutional medical FL still incurs prohibitive bandwidth~\cite{haripriya2025privacy}, and communication-efficient extensions such as knowledge distillation~\cite{wu2022communicationkd} and metaheuristic aggregation~\cite{abdolmaleki2025syncgwo} treat compression and topology choice as separate problems. We label this open problem \textbf{Q4} \textit{(communication overhead, even in P2P)} and answer it through the additive head-personalization, bfloat16-quantization, and tunable active-set cardinality {\small $\kappa$}, optimized jointly on the accuracy-bandwidth Pareto frontier (Sec.~\ref{sec:headline}, with the row-by-row ablation in Sec.~\ref{sec:ablations}).

\vspace{-4.5mm}
\section{Problem Formulation}\label{sec:problem_formulation}
\vspace{-3.5mm}

\textbf{Goal.} We seek to learn, for each of {\small $n$} federated centers, a sequence of local predictors that \emph{maximizes predictive accuracy} on a center's evolving data distribution while \emph{minimizing the cumulative communication overhead} of inter-center collaboration, \emph{under continuous concept drift} in patient demographics, treatment protocols, and equipment. \textbf{Predictive accuracy} is task-dependent and enters the framework through a per-center utility functional {\small $U_i \in [0,1]$. $U_i$} is the per-center Area Under the Receiver Operating Characteristic curve (AUROC) for binary in-hospital mortality classification.\\
At each round {\small $t \in \{1,\dots,R\}$}, every center {\small $i \in \{1,\dots,n\}$} holds local data drawn from an unknown, time-varying joint distribution {\small $P^{(i)}_t(X,Y)$} shaped by the center's patient demographics, comorbidity profile, and care protocols, observed only through patient-level samples, fits local parameters {\small $\mathbf{w}_{i,t} \in \mathbb{R}^{|W|}$}, and may exchange them with a per-round active peer set {\small $\mathcal{P}_{i,t} \subseteq \{1,\dots,n\}\setminus\{i\}$}. We write {\small $A_i(\mathbf{w}_{i,t}) := \mathbb{E}_{(x,y)\sim P^{(i)}_t}\!\left[\mathrm{AUROC}(\mathbf{w}_{i,t};x,y)\right]$} for the expected accuracy of {\small $i$}'s model and treat {\small $\mathcal{P}_{i,t}=\emptyset$} as a legal first-class action (\emph{intentional rest}, formalized in Lemma~\ref{lem:isolation}). \textsc{ABPS} jointly chooses, for every center and every round, the active peer set and the local update rule, solving
{\footnotesize
\vspace{-2mm}
\begin{equation}\label{eq:objective}
\max_{\{\mathcal{P}_{i,t},\,\mathbf{w}_{i,t}\}}\;
\underbrace{\frac{1}{n}\sum_{i=1}^{n}\frac{1}{R}\sum_{t=1}^{R} A_i(\mathbf{w}_{i,t})}_{\text{predictive accuracy}}
\;\;\;\text{subject to}\;\;\;
\underbrace{C_{\mathrm{total}} \le B}_{\text{bandwidth budget}},\;\;
\underbrace{\textstyle\sup_{m,k}\, D\!\left[P^{(i)}_{m,k}\,\|\,P^{(i)}_{m,k-1}\right] \le \epsilon_m}_{\text{drift bound at scale }m}.
\end{equation}
}
The maximand is the across-center, across-round mean accuracy. {\small $C_{\mathrm{total}} \le B$} caps cumulative communication (Eq.~\ref{eq:cost} below). The drift constraint bounds the per-window divergence by a scale-specific tolerance {\small $\epsilon_m$} at each temporal scale {\small $m \in \{1,2,3\}$}. When a realized shift exceeds {\small $\epsilon_m$}, A3 (Sec.~\ref{sec:theory}) breaks at scale {\small $m$}, the Bayesian posterior decays toward the prior, the UCB rankings re-order, and the propose-reject mechanism re-forms {\small $\mathcal{P}_{i,t+1}$}. Equation~\ref{eq:objective} thus makes the three competing forces explicit: accuracy as the maximand, bandwidth as the budget, and drift as the constraint.\\
\textbf{Communication cost.} Let {\small $\mathcal{P}_t = \{(i,j) : j \in \mathcal{P}_{i,t}\}$} denote the active peer-pair set in round {\small $t$}, {\small $|W_i|$} the size of center {\small $i$}'s parameter vector, and {\small $c_{\mathrm{egress}}$} the per-GB egress cost (e.g., {\small $\$0.09$}/GB on AWS~\cite{aws2023pricing}). It is worth mentioning here that continuously broadcasting parameters across {\small $100{,}000{+}$} global healthcare centers typically incurs a prohibitive bandwidth footprint~\cite{mcmahan2017communication}, and recent medical-FL benchmarks~\cite{haripriya2025privacy} confirm that one {\small $50$}-round sweep of a VGG-16-class model exceeds {\small $276{,}000$} MB of cross-client traffic. Distillation~\cite{wu2022communicationkd} and bandwidth-aware aggregation~\cite{abdolmaleki2025syncgwo} cut these figures, but neither couples compression to peer choice. The cumulative cost across {\small $R$} rounds is
{\footnotesize
\vspace{-2mm}
\begin{equation}\label{eq:cost}
    C_{\mathrm{total}} \;=\; c_{\mathrm{egress}} \sum_{t=1}^{R} \sum_{(i,j) \in \mathcal{P}_t} \bigl(|W_i| + |W_j|\bigr).
\end{equation}}
\noindent In Eq.~\ref{eq:cost}, the summand accounts for the bidirectional transfer of parameter aggregation. Under the uniform-architecture specialization {\small $|W_i|{=}|W|$}, this reduces to {\small $C_{\mathrm{total}} = 2\, c_{\mathrm{egress}}\, |W|\, \sum_t |\mathcal{P}_t|$}, the form used in Theorem~\ref{thm:regret} and Lemma~\ref{lem:isolation}.

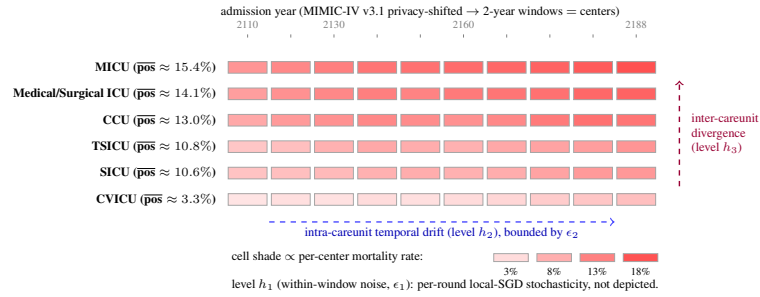
\begin{wrapfigure}{r}{0.72\textwidth}
\vspace{-4mm}
\centering
\small
\resizebox{0.72\columnwidth}{!}{%
\begin{tikzpicture}[
    x=0.85cm, y=0.52cm,
    every node/.style={font=\scriptsize},
    cellbox/.style={draw=black!35, line width=0.25pt},
    unitlabel/.style={anchor=east, font=\scriptsize\bfseries},
    driftarrow/.style={->, thick, blue!70, dashed},
    divarrow/.style={->, thick, purple!80!black, dashed},
]
\foreach \y/\name/\op in {%
  0/CVICU ($\overline{\text{pos}}\approx3.3\%$)/{0.12, 0.14, 0.13, 0.15, 0.14, 0.16, 0.18, 0.22, 0.25, 0.28},%
  1/SICU ($\overline{\text{pos}}\approx10.6\%$)/{0.26, 0.28, 0.30, 0.34, 0.36, 0.34, 0.36, 0.40, 0.44, 0.45},%
  2/TSICU ($\overline{\text{pos}}\approx10.8\%$)/{0.25, 0.28, 0.34, 0.38, 0.36, 0.34, 0.38, 0.40, 0.42, 0.46},%
  3/CCU ($\overline{\text{pos}}\approx13.0\%$)/{0.38, 0.44, 0.48, 0.52, 0.50, 0.48, 0.52, 0.58, 0.62, 0.60},%
  4/{Medical/Surgical ICU ($\overline{\text{pos}}\approx14.1\%$)}/{0.42, 0.48, 0.52, 0.56, 0.54, 0.58, 0.60, 0.62, 0.66, 0.68},%
  5/MICU ($\overline{\text{pos}}\approx15.4\%$)/{0.46, 0.52, 0.56, 0.60, 0.62, 0.64, 0.66, 0.68, 0.72, 0.76}%
} {
  \node[unitlabel] at (-0.1, \y) {\name};
  \foreach \c [count=\cc from 0] in \op {
    \fill[red!90, opacity=\c] (\cc+0.05, \y-0.22) rectangle (\cc+0.95, \y+0.22);
    \draw[cellbox] (\cc+0.05, \y-0.22) rectangle (\cc+0.95, \y+0.22);
  }
}
\foreach \x in {0,1,...,9} {\draw[gray] (\x+0.5, 6.25) -- +(0, -0.12);}
\foreach \x/\lbl in {0/2110, 2/2130, 5/2160, 9/2188} {
  \node[above, font=\tiny, gray] at (\x+0.5, 6.27) {\lbl};
}
\node[above, font=\scriptsize] at (4.5, 6.65)
  {admission year (MIMIC-IV v3.1 privacy-shifted $\to$ 2-year windows $=$ centers)};
\draw[driftarrow] (1.0, -0.85) -- (9.0, -0.85);
\node[below, font=\scriptsize, text=blue!70!black] at (5.0, -0.85)
  {intra-careunit temporal drift (level $h_2$), bounded by $\epsilon_2$};
\draw[divarrow] (10.5, 0.5) -- (10.5, 4.5);
\node[right, font=\scriptsize, text=purple!80!black, align=left] at (10.65, 2.5)
  {inter-careunit\\divergence\\(level $h_3$)};
\node[font=\scriptsize, anchor=west] at (0, -2.2)
  {cell shade $\propto$ per-center mortality rate:};
\foreach \x/\op/\lbl in {6.2/0.15/3\%, 7.2/0.35/8\%, 8.2/0.55/13\%, 9.2/0.75/18\%} {
  \fill[red!90, opacity=\op] (\x, -2.35) rectangle (\x+0.8, -2.05);
  \draw[cellbox] (\x, -2.35) rectangle (\x+0.8, -2.05);
  \node[font=\tiny, below] at (\x+0.4, -2.35) {\lbl};
}
\node[font=\scriptsize, anchor=west] at (0, -3.25)
  {level $h_1$ (within-window noise, $\epsilon_1$): per-round local-SGD stochasticity, not depicted.};
\end{tikzpicture}
}
\vspace{-3mm}
\caption{\scriptsize Hierarchical temporal-shift model, illustrated on MIMIC-IV v3.1 as one concrete instantiation. The construction is dataset-agnostic and applies to any longitudinal multi-center cohort. In this instantiation each cell is one of {\small $n{=}230$} federated centers (careunit {\small $\times$} 2-year window). Cell shade encodes in-hospital mortality rate, and rows sort six careunits by mean mortality. Horizontal variation within a row illustrates the level-{\small $h_2$} within-group term bounded by {\small $\epsilon_2$}; because the windows are shifted-year buckets rather than calendar years, on this axis that variation is of the order of sampling variation (Sec.~\ref{sec:experimental}). Vertical shade jumps show level-{\small $h_3$} inter-group divergence that goal-aware pre-filter (Sec.~\ref{sec:goal_filter}) reasons about without exchanging model parameters.}
\label{fig:hierarchical_drift}
\vspace{-4.75mm}
\end{wrapfigure}

\textbf{Hierarchical concept drift.} The drift constraint of Eq.~\ref{eq:objective} bounds, at three temporal scales {\small $m \in \{1,2,3\}$} (short-term operational, medium-term seasonal, long-term demographic), the inter-window divergence {\small $D[P^{(i)}_{m,k} \,\|\, P^{(i)}_{m,k-1}] \le \epsilon_m$} between consecutive windows of length {\small $\Delta t_m$} **(Figure~\ref{fig:hierarchical_drift})**~\cite{gama2014survey, rahimli2024federated, cuturi2017soft, dusing2025distribution, rahad2025kl-feddis}. The window size {\small $\Delta t_m$} admits an online empirical estimator, e.g., a Welch {\small $t$}-test that expands the window while {\small $|\mu_t-\mu_{t+k}|/\sqrt{\sigma_t^2/N_t + \sigma_{t+k}^2/N_{t+k}}$} stays below {\small $t_{\mathrm{crit}}$} at {\small $\alpha{=}0.05$}. (Note that during experimental validation, we fix {\small $\Delta t_2{=}2$} years and leave adaptive estimation of {\small $\{\Delta t_m\}$} to future work, since online window adaptation is orthogonal to the partner-selection contribution we focus on here, and Theorem~\ref{thm:convergence} already bounds the within-window cost explicitly through {\small $\epsilon_m$}, so the headline accuracy-bandwidth claim is unaffected by any reasonable choice of {\small $\Delta t_2$}.)

\vspace{-4mm}
\section{Methodology: Adaptive Bayesian Partner Selection}\label{sec:methodology}

\vspace{-4mm}
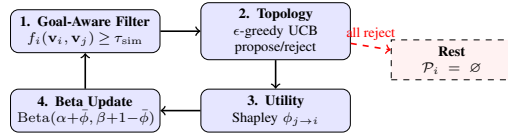
\begin{wrapfigure}{r}{0.48\textwidth}
\vspace{-5mm}
\small
\centering
\resizebox{0.48\columnwidth}{!}{%
\begin{tikzpicture}[
    node distance=0.5cm and 0.7cm,
    block/.style={rectangle, draw, fill=blue!10, text width=2.4cm, align=center, rounded corners, minimum height=0.75cm, font=\scriptsize},
    arrow/.style={->, thick},
    label/.style={font=\scriptsize\itshape, text=black!70}
]
\node[block] (metadata) {\textbf{1. Goal-Aware Filter}\\$f_i(\mathbf{v}_i,\mathbf{v}_j)\!\ge\!\tau_{\mathrm{sim}}$};
\node[block] (ucb) [right=of metadata] {\textbf{2. Topology}\\$\epsilon$-greedy UCB propose/reject};
\node[block] (shapley) [below=of ucb] {\textbf{3. Utility}\\Shapley $\phi_{j\to i}$};
\node[block] (bayes) [left=of shapley] {\textbf{4. Beta Update}\\$\mathrm{Beta}(\alpha{+}\bar\phi,\beta{+}1{-}\bar\phi)$};
\draw[arrow] (metadata) -- (ucb);
\draw[arrow] (ucb) -- (shapley);
\draw[arrow] (shapley) -- (bayes);
\draw[arrow] (bayes) -- (metadata);
\node[rectangle, draw, dashed, fill=red!5, text width=1.9cm, align=center, font=\scriptsize, right=of ucb, yshift=-0.5cm] (isolate) {\textbf{Rest}\\$\mathcal{P}_i=\varnothing$};
\draw[->, dashed, thick, red] (ucb) -- node[above, font=\scriptsize, text=red] {all reject} (isolate);
\end{tikzpicture}
}
\vspace{-4mm}
\caption{\scriptsize \textsc{ABPS} round-level pipeline. Centers prune candidates via a goal-aware metadata filter (Eq.~\ref{eq:goal_score}), propose to peers ranked by UCB on Beta belief, observe Shapley marginal utility, and update the belief. When all proposals are rejected, the center rests with $\mathcal{P}_i = \varnothing$ (Lemma~\ref{lem:isolation}).}
\label{fig:algorithm_overview}
\vspace{-4mm}
\end{wrapfigure}

\textsc{ABPS} models inter-center collaboration as a probabilistic, temporally adaptive process built from Bayesian-core components, a goal-aware metadata pre-filter (Sec.~\ref{sec:filter}), a Beta-Bernoulli posterior over each peer's marginal utility (Sec.~\ref{sec:belief}), an {\small $\epsilon$}-greedy UCB propose-reject topology rule (Sec.~\ref{sec:topology_subsec}), and a \emph{rest} action when no proposal is accepted, and three communication-efficiency extensions that compose additively on top of the core, namely, head personalization, bfloat16 quantization, and tunable {\small $\kappa$}, are all discussed hereafter and analyzed in Sec.~\ref{sec:headline}). Figure~\ref{fig:algorithm_overview} summarizes the round-level pipeline, and the full pseudocode is given in Algorithm~\ref{alg:adaptive_partner_selection} in Appendix~\ref{app:algorithm}.

\vspace{-4mm}
\subsection{Goal-Aware Metadata Pre-Filter}\label{sec:filter}\label{sec:goal_filter}
\vspace{-3mm}
\emph{To address \textbf{Q1} on goal-heterogeneous collaboration from Sec.~\ref{sec:related_work},} before invoking the expensive Shapley-UCB stage, each center {\small $i$} computes a {\small $d$}-dimensional summary metadata vector {\small $\mathbf{v}_i \in \mathbb{R}^d$} of non-sensitive aggregates (positive-class rate, {\small $\log n_{\mathrm{train}}$}, normalized mean year of admission, and for goal-aware variants a {\small $12$}-dim Charlson-comorbidity prevalence). Every entry is a cohort-level scalar, not a per-patient feature, so {\small $\mathbf{v}_i$} carries no Protected Health Information. No model parameters cross the network at this stage, consistent with HIPAA and inter-institutional data-sharing constraints.\\
\textbf{Goal-aware admission rule.} Real healthcare federations are \emph{not} uniformly similarity-seeking: the three institution types in Sec.~\ref{sec:intro} have distinct objectives. A community hospital prefers \emph{homogeneity}, collaborating with demographically similar peers to improve transfer learning for under-represented cohorts. An academic medical center seeks \emph{diversity}, leveraging exposure to heterogeneous case-mix. In contrast, an Integrated Delivery Network emphasizes \emph{alignment}, selecting peers with matching target KPIs (e.g., mortality rates) to meet system-level objectives. We generalize the admission rule into a per-center goal-scoring function {\small $f_i : \mathbb{R}^d \times \mathbb{R}^d \to \mathbb{R}$}:
{\footnotesize
\vspace{-2.5mm}
\begin{equation}\label{eq:goal_score}
    f_i(\mathbf{v}_i, \mathbf{v}_j) \;=\;
    \begin{cases}
        \cos(\mathbf{v}_i, \mathbf{v}_j), & \text{goal} = \mathrm{homogeneity}\\
        -\cos(\mathbf{v}_i, \mathbf{v}_j), & \text{goal} = \mathrm{diversity}\\
        -|v_{i,0} - v_{j,0}|, & \text{goal} = \mathrm{alignment}
    \end{cases}
\end{equation}
}
\noindent In the above equation, {\small $v_{\cdot,0}$} is the pos-rate coordinate used as the KPI marker. Peer {\small $j$} is admitted to {\small $\mathcal{C}_i$} iff {\small $f_i(\mathbf{v}_i,\mathbf{v}_j) \ge \tau_{\mathrm{sim}}$}. Each formulation lives on a different scale, so {\small $\tau_{\mathrm{sim}}$} is auto-calibrated per goal to retain a fixed fraction (we use {\small $25\%$}) of candidate pairs: this makes goal choices directly comparable. The threshold can equivalently be learned online via an exponential moving average (EMA) on the acceptance rate. The static calibration suffices for our experiments.\\
\textbf{Complexity and theoretical preservation.} The pre-filter is {\small $\mathcal{O}(nd)$} per round and transmits only {\small $\mathbf{v}_i$} (a few dozen bytes) per center. All downstream Shapley evaluations operate on the filtered arm set, reducing the expected per-round bandwidth by the same fraction. Theorems~\ref{thm:convergence} and \ref{thm:regret} transfer verbatim once the arm set is restricted to {\small $\mathcal{C}_i$}. The only added bias is whether the true best peer survives the filter, which we bound in Appendix~\ref{app:proofs} via a standard top-{\small $k$} coverage argument for descriptor-based recall. Empirically (Sec.~\ref{sec:headline}), the diversity goal activates Lemma~\ref{lem:isolation}'s {\small $|\mathcal{P}_i|=0$} regime in {\small $\sim 40\%$} of rounds on the biomedical dataset, the first empirical observation of intentional isolation in our study.

\vspace{-4mm}
\subsection{Bayesian Belief Updating on Marginal Utility}\label{sec:belief}
\vspace{-3mm}
Rather than maintaining a belief over a peer's explicitly shared raw parameters, center {\small $i$} evaluates the actual \emph{marginal utility} (performance gain) a peer {\small $j$} provides. We model this utility using the Shapley value to objectively allocate credit across historical collaboration subsets {\small $\mathcal{S} \subseteq \mathcal{P}_i \setminus \{j\}$}:
{\footnotesize
\vspace{-2mm}
\begin{equation}\label{eq:shapley}
    \phi_{j \to i} = \sum_{\mathcal{S} \subseteq \mathcal{P}_i \setminus \{j\}} \frac{|\mathcal{S}|! \, (|\mathcal{P}_i| - |\mathcal{S}| - 1)!}{|\mathcal{P}_i|!} \left( U_i(\mathcal{S} \cup \{j\}) - U_i(\mathcal{S}) \right),
\end{equation}
}
In the above equation, {\small $U_i(\cdot) \in [0,1]$} is the local model's normalized held-out validation performance (we use AUROC). Because {\small $U_i$} is bounded in {\small $[0,1]$}, the per-peer Shapley contribution {\small $\phi_{j \to i}$} is bounded in {\small $[-1,+1]$}. We map it to the unit interval via the affine clip

{\footnotesize
\vspace{-3mm}
\begin{equation}\label{eq:phi_bar}
    \bar{\phi}_{j \to i} \;=\; \mathrm{clip}_{[0,1]} \!\left(\frac{\phi_{j \to i} - \phi_{\min}}{\phi_{\max} - \phi_{\min}}\right),
\end{equation}}
\noindent with truncation thresholds {\small $\phi_{\min}, \phi_{\max}$} chosen so that a single anomalous round cannot saturate the posterior (we use {\small $[-0.1,+0.1]$} throughout this paper). Modeling {\small $\bar{\phi}_{j \to i}$} as a soft Bernoulli observation, the conjugate Beta prior {\small $\bar{\phi}_{j \to i} \sim \mathrm{Beta}(\alpha_{i \to j}, \beta_{i \to j})$} admits the closed-form update
{\footnotesize
\begin{equation}\label{eq:beta_update}
    \alpha_{i \to j}^{(k+1)} \gets \alpha_{i \to j}^{(k)} + \bar{\phi}_{j \to i}, \qquad
    \beta_{i \to j}^{(k+1)}  \gets \beta_{i \to j}^{(k)}  + (1 - \bar{\phi}_{j \to i}).
\end{equation}
}
\noindent The Beta-Bernoulli formulation offers three properties. (i)~The posterior support {\small $[0,1]$} matches the bounded co-domain of {\small $\bar{\phi}_{j \to i}$}, ruling out the model mis-specification that would otherwise arise when Shapley contributions are negative. (ii)~The posterior variance {\small $\alpha\beta/[(\alpha{+}\beta)^2(\alpha{+}\beta{+}1)]$} admits the Hoeffding-style concentration exploited in Theorem~\ref{thm:convergence}. (iii)~The upper-confidence bound(in Sec.~\ref{sec:topology_subsec}) for partner ranking takes the standard UCB1 form and inherits sublinear regret (Theorem~\ref{thm:regret}).

\vspace{-4mm}
\subsection{Collaborative Topology Formulation}\label{sec:topology_subsec}
\vspace{-3mm}
\emph{Recall \textbf{Q2} (on selective collaboration under heterogeneity) and \textbf{Q3} (on adaptive isolation under drift) from Sec.~\ref{sec:related_work}.} The propose-reject protocol below operationalizes both, with the explicit \emph{rest} action carrying through to the optimality result of Lemma~\ref{lem:isolation}. The global time horizon {\small $T$} is partitioned into discrete interaction intervals {\small $\{t_1,\dots,t_K\}$}. Collaboration decisions among the centers are governed by a propose-and-reject protocol approximating a stable-marriage solution, augmented with an {\small $\epsilon$}-greedy Upper Confidence Bound (UCB) strategy~\cite{auer2002finite}. During each period {\small $t_k$}, center {\small $i$} computes a utility-based UCB score for every candidate peer {\small $j$}:
{\footnotesize
\vspace{-3mm}
\begin{equation}\label{eq:ucb}
    \mathrm{UCB}_{j \to i}(t_k) \;=\; \hat{\mu}_{i \to j}(t_k) \;+\; \gamma \sqrt{\frac{2 \log t_k}{n_{i \to j}(t_k) + 1}},
\end{equation}
}
Here, {\small $\hat{\mu}_{i \to j}(t_k) = \alpha_{i \to j}^{(t_k)} / \big(\alpha_{i \to j}^{(t_k)} + \beta_{i \to j}^{(t_k)}\big)$} is the posterior mean of the Beta belief~\eqref{eq:beta_update}, {\small $n_{i \to j}(t_k)$} is the number of past observations from {\small $j$}, and {\small $\gamma > 0$} is the exploration weight. With probability {\small $1-\epsilon$}, center {\small $i$} sequentially proposes collaboration to peers sorted by descending {\small $\mathrm{UCB}_{j \to i}$}. With probability {\small $\epsilon$}, it purposefully queries a uniformly random unproven neighbor. A receiving peer {\small $j$} evaluates the proposal and accepts only if its own UCB on {\small $i$} exceeds the threshold {\small $\tau_{\mathrm{acc}}$}. If {\small $j$} rejects, {\small $i$} extends the proposal to the next highest-ranked peer until either {\small $\kappa$} acceptances accrue or the candidate list is exhausted. Critically, if all proposals are rejected, center {\small $i$} \emph{rests}: its active peer set {\small $\mathcal{P}_i$} collapses to {\small $\varnothing$} and no parameters are exchanged this round. This isolation mechanism is a deliberate design feature, not a system failure. In highly heterogeneous networks, forcing connections with dissimilar nodes degrades local performance via negative transfer. Choosing isolation whenever the expected gain falls below the threshold helps protect local utility and conserves bandwidth (Lemma~\ref{lem:isolation}). Resting is per-round and self-correcting. The next round re-runs the propose-reject loop with an updated posterior and a UCB exploration radius that grows in $t$, and the $\epsilon$-greedy swap admits a uniformly-random peer with probability $\epsilon$ regardless of UCB ranking, so collaboration resumes the moment any peer's UCB rises above $\tau_{\mathrm{acc}}$ or drift makes a previously-low-utility peer informative again.

\vspace{-5mm}
\section{Theoretical Guarantees and Complexity Analysis}\label{sec:theory}
\vspace{-3.5mm}
We analyze \textsc{ABPS} along posterior concentration (Theorem~\ref{thm:convergence}), partner-selection regret (Theorem~\ref{thm:regret}), and Bayes-optimality of isolation (Lemma~\ref{lem:isolation}). Proofs are in Appendix~\ref{app:proofs}. The analysis rests on three assumptions. \textbf{A1} bounded utility, {\small $U_i \in [0,1]$}, so {\small $\bar\phi_{j\to i} \in [0,1]$} via the affine clip of Eq.~\ref{eq:phi_bar} (see Sec.~\ref{sec:belief}). \textbf{A2} conditional independence of the {\small $\bar\phi$} observations given {\small $\mu^\star_{i\to j}$}, justified by independent local stochastic gradient descent draws and disjoint mini-batches per round. \textbf{A3} quasi-stationary drift {\small $|\mathbb{E}[\bar\phi^{(k)}_{j\to i}] - \mu^\star_{i\to j}| \le \epsilon_m$} within any window {\small $\Delta t_m$}, binding the theory to the hierarchical formulation of Sec.~\ref{sec:problem_formulation}. For the filtered-arm-set refinement of Theorem~\ref{thm:regret}, we additionally assume \textbf{A4} that the metadata descriptor {\small $\mathbf{v}_i$} is {\small $L$}-Lipschitz informative about utility, {\small $|\mu^\star_{i \to j} - \mu^\star_{i \to j'}| \le L\,\|\mathbf{v}_j - \mathbf{v}_{j'}\|$} (see Appendix~\ref{app:proof_regret} for the precise statement and use).
\vspace{-3mm}
\subsection{Convergence of the Beta Posterior}
\vspace{-3mm}

\begin{theorem}[Beta-belief concentration]\label{thm:convergence}
Let {\small $\bar{\phi}_{j \to i}^{(1)}, \dots, \bar{\phi}_{j \to i}^{(K)} \in [0,1]$} be the clipped Shapley observations collected by center {\small $i$} for peer {\small $j$} across {\small $K$} rounds inside a single drift window, with empirical mean {\small $\bar{\mu}_K = \frac{1}{K} \sum_{k=1}^K \bar{\phi}_{j \to i}^{(k)}$} and posterior mean {\small $\hat{\mu}_K = \frac{\alpha_0 + K \bar{\mu}_K}{\alpha_0 + \beta_0 + K}$}. Under A1-A3, for any {\small $\delta \in (0,1)$} with probability at least {\small $1-\delta$}
{\footnotesize
\vspace{-3mm}
\begin{equation}\label{eq:concentration}
    \big| \hat{\mu}_K - \mu^\star_{i \to j} \big| \;\le\; \underbrace{\sqrt{\frac{\log(2/\delta)}{2K}}}_{\text{Hoeffding}} \;+\; \underbrace{\frac{\alpha_0 + \beta_0}{\alpha_0 + \beta_0 + K}}_{\text{prior decay}} \;+\; \underbrace{\epsilon_m \vphantom{\sqrt{\frac{\log(2/\delta)}{2K}}}}_{\text{drift}}.
\end{equation}
}
Hence {\small $\hat{\mu}_K \xrightarrow{p} \mu^\star_{i \to j}$} as {\small $K \to \infty$} with {\small $\epsilon_m \to 0$}.
\end{theorem}
\vspace{-3mm}
The proof (Appendix~\ref{app:proof_convergence}) decomposes the error into Hoeffding noise ({\small $\mathcal{O}(1/\sqrt{K})$}), prior decay ({\small $\mathcal{O}(1/K)$}), and within-window drift bounded by {\small $\epsilon_m$}, the last of which motivates the Welch-{\small $t$} window selection of Sec.~\ref{sec:problem_formulation}.

\vspace{-4mm}
\subsection{Regret of the UCB Partner Selection Bandit}
\vspace{-3mm}
We bound the regret of the per-center partner-selection problem viewed as a {\small $\kappa$}-armed bandit\cite{lattimore2020bandit} over the candidate pool {\small $\mathcal{C}_i$}. Let {\small $\mu^\star_{(1)} \ge \mu^\star_{(2)} \ge \dots$} denote the ordered true utilities of {\small $i$}'s candidate peers and define the suboptimality gaps {\small $\Delta_j = \mu^\star_{(1)} - \mu^\star_{(j)}$}.
\begin{theorem}[Sublinear partner-selection regret]\label{thm:regret}
Under assumptions A1-A2 and a stationary window ({\small $\epsilon_m = 0$}), the cumulative regret of \textsc{ABPS} with exploration parameter {\small $\gamma = \sqrt{2}$} and {\small $\epsilon$}-greedy exploration probability {\small $\epsilon \in [0,1)$} over {\small $T$} rounds satisfies
{\footnotesize
\begin{equation}\label{eq:regret}
    \mathcal{R}(T) \;\le\; \kappa \sum_{j: \Delta_j > 0} \!\left( \frac{8 \log T}{\Delta_j} + \Delta_j \!\left(1 + \frac{\pi^2}{3}\right)\right) \;+\; \epsilon \, T \cdot \mathbb{E}_{j \sim \mathrm{Unif}(\mathcal{C}_i)}[\Delta_j].
\end{equation}
}
Setting {\small $\epsilon = \mathcal{O}(1/\sqrt{T})$} recovers the standard {\small $\mathcal{O}(\log T)$} regret rate of UCB1 up to the factor {\small $\kappa$}.
\end{theorem}
\vspace{-3mm}
The proof (Appendix~\ref{app:proof_regret}) bounds the greedy phase via canonical UCB1 analysis~\cite{auer2002finite} scaled by {\small $\kappa$} and the exploratory phase via the {\small $\epsilon T \cdot \mathbb{E}[\Delta_j]$} term, with a filtered-arm-set refinement under A4 that improves the constants when the goal-aware pre-filter is active. In stationary windows, \textsc{ABPS} therefore matches an oracle that always proposes to the {\small $\kappa$} best peers.

\vspace{-4mm}
\subsection{Optimality of Intentional Isolation}\label{sec:isolation_lemma}
\vspace{-2mm}

\emph{Recall \textbf{Q3} (adaptive isolation under drift) from Sec.~\ref{sec:related_work}.} The lemma below is the formal optimality guarantee that the rest of the action of Sec.~\ref{sec:topology_subsec} promised.
\begin{lemma}[Intentional isolation dominates forced collaboration]\label{lem:isolation}
Let {\small $V_i^{\mathrm{rest}}$} denote the expected one-step utility of center {\small $i$} when it rests ({\small $\mathcal{P}_i = \varnothing$}) and {\small $V_i^{\mathrm{coll}}(\mathcal{P})$} the expected utility when it forcibly collaborates with peer set {\small $\mathcal{P} \neq \varnothing$}. Suppose the per-peer expected marginal utility satisfies {\small $\mu^\star_{i \to j} < c_{\mathrm{neg}}$} for every {\small $j \in \mathcal{P}$}, where {\small $c_{\mathrm{neg}}$} is the negative-transfer threshold defined by {\small $V_i^{\mathrm{coll}}(\{j\}) = V_i^{\mathrm{rest}}$} when {\small $\mu^\star_{i \to j} = c_{\mathrm{neg}}$}. Then
{\small
\vspace{-2mm}
\begin{equation}\label{eq:isolation}
    V_i^{\mathrm{rest}} \;>\; V_i^{\mathrm{coll}}(\mathcal{P}),
\end{equation}
}
and the bandwidth saved by resting is exactly {\small $C_i^{\mathrm{rest}} = 2|\mathcal{P}| \cdot c_{\mathrm{egress}} \cdot |W|$} per round (cf.\ eq.~(1)).
\end{lemma}
\vspace{-3mm}
The proof (Appendix~\ref{app:proof_isolation}) combines Shapley efficiency with the affine clip of Eq.~\ref{eq:phi_bar}, which fixes {\small $c_{\mathrm{neg}} = -\phi_{\min}/(\phi_{\max}-\phi_{\min}) = 0.5$} for {\small $(\phi_{\min},\phi_{\max}) = (-0.1,+0.1)$}, exactly {\small $\tau_{\mathrm{acc}}$}. \textsc{ABPS} approximates this oracle by resting whenever every UCB falls below {\small $\tau_{\mathrm{acc}}$}, and Theorem~\ref{thm:convergence} guarantees convergence to the oracle as {\small $K \to \infty$}.
\vspace{-4mm}

\subsection{Computational Complexity}\label{sec:complexity}
\vspace{-3mm}

Let {\small $\rho \in (0, 1]$} denote the expected top-{\small $k$} coverage of the goal-aware pre-filter (Sec.~\ref{sec:goal_filter}), i.e.\ the fraction of peers that survive the {\small $\tau_{\mathrm{sim}}$} threshold. When {\small $\tau_{\mathrm{sim}}$} is calibrated to retain a prescribed keep-fraction, {\small $\rho$} equals that target (we use {\small $\rho = 0.25$}). Define the post-filter candidate size {\small $|\mathcal{C}_i| \approx \rho (n-1)$}.\\
The per-round computational cost at each center is strictly bounded and decomposes into four components. (1) \textit{\underline{Goal-aware filter}:} Evaluating {\small $f_i(\mathbf{v}_i, \mathbf{v}_j)$} for each of the {\small $n-1$} peers using {\small $d$}-dimensional metadata incurs a cost of {\small $\mathcal{O}(n d)$}, where {\small $d \le 15$} in our implementation (3 base features and 12 comorbidity indicators); (2) \textit{\underline{UCB ranking and propose--reject}:} Ranking candidates in the filtered set {\small $|\mathcal{C}_i| \approx \rho n$} and traversing the ordered list requires {\small $\mathcal{O}(\rho n \log(\rho n))$} time requires a {\small $\rho$}-factor reduction compared to the unfiltered case; (3) \textit{\underline{Shapley credit assignment}:} The exact computation of this involves evaluating {\small $2^{|\mathcal{P}_i|}$} coalitions. Since {\small $|\mathcal{P}_i| \le \kappa$} by design, the cost is {\small $\mathcal{O}(2^\kappa)$}, independent of {\small $n$}. For {\small $\kappa > 5$}, we instead employ Truncated Monte Carlo (TMC) Shapley~\cite{ghorbani2019data}, which achieves an {\small $\varepsilon$}-accurate estimate in {\small $\mathcal{O}(\kappa \log \kappa / \varepsilon^2)$} samples; and (4) \textit{\underline{Beta--Bernoulli update}:} Posterior updates require only closed-form scalar operations, yielding {\small $\mathcal{O}(1)$} complexity.\\
The total per-round computation at center {\small $i$} is therefore {\small $\mathcal{O}(nd + \rho n \log(\rho n) + 2^\kappa)$}. For the default {\small $\rho = 0.25$} and {\small $\kappa = 1$} this collapses to {\small $\mathcal{O}(nd)$}, \emph{independent of the number of model parameters {\small $|W|$}}.\\
\textbf{Communication.} Each round transmits (i)~the metadata vector {\small $\mathbf{v}_i$} once at filter time (a few dozen bytes per center, \emph{no model weights}) and (ii)~the active-peer aggregation for the {\small $\kappa$} peers that survive both the filter and the UCB threshold, costing {\small $2 \kappa |W|$} per active center per round. With personalization (Sec.~\ref{sec:headline}), only the shared trunk is transmitted. With quantization, each {\small $|W|$} is reduced to its bfloat16 footprint. When {\small $|\mathcal{P}_i| = 0$} (intentional isolation, empirically realized under the diversity goal), the communication collapses to zero for that center-round.
\vspace{-5mm}

\section{Experimental Evaluation}\label{sec:experimental}
\vspace{-4mm}
We evaluate \textsc{ABPS} along four axes: (i)~predictive accuracy under realistic non-IID partitioning of MIMIC-IV, a publicly available database sourced from the Beth Israel Deaconess Medical Center (BIDMC) electronic health record~\cite{johnson2023mimic}, (ii)~cumulative transmitted bytes as a direct proxy for communication cost, (iii)~the empirical realization of the intentional-isolation property of Lemma~\ref{lem:isolation}, and (iv)~an ablation decomposing the contribution of each framework extension. Code, sbatch scripts, and per-seed per-method result JSONs will accompany the camera-ready submission.
\vspace{-4mm}
\subsection{Dataset and Federation Setup}
\vspace{-3mm}
The task at hand is the binary in-hospital mortality prediction over the first 24 hours of an ICU stay~\cite{johnson2023mimic}. After applying the standard age filter ({\small $18 \le \text{age} \le 95$}), the cohort contains roughly {\small $76{,}000$} ICU stays. The patient features: demographics (age, gender, race), admission context (admission type, location, insurance), Charlson-style comorbidity binaries, discussed in Sec. \ref{sec:metrics}, derived from ICD-10/ICD-9 codes, and aggregated first-24h vitals (heart rate, systolic/diastolic/mean BP, respiration rate, SpO\textsubscript{2}, temperature) summarized as {\small $\{\mathrm{mean}, \min, \max\}$}. Continuous features are standardized \emph{per center} on the local training split to respect federated isolation.\\
Our experiments rely on a \textit{careunit-by-year} partitioning of MIMIC-IV, where a center is defined as the Cartesian product of a care unit and a consecutive two-year admission window. The care units include CVICU, CCU, MICU, Medical/Surgical ICU, SICU, and TSICU, yielding {\small $n=230$} centers spanning the shifted temporal range 2110 to 2191 in MIMIC-IV v3.1. MIMIC-IV de-identifies dates by a single random offset per subject, applied uniformly to all of that subject's events~\cite{johnson2023mimic}, so within-patient order is exact, but the shifted-year windows are not aligned with calendar time: across the 71{,}008 stays in the partition, Cram\'er's $V$ between the assigned window and the published \texttt{anchor\_year\_group} field is 0.045, and mean window purity is 0.335 against a chance value of 0.336. We therefore treat the windows as a partitioning device rather than a calendar axis, and the empirical evidence of drift over calendar time in Appendix~\ref{app:drift_evidence} uses the published era field. What the construction provides is a deterministic partition into many small centers whose outcome distributions differ: the mean pairwise Jensen-Shannon divergence between the Bernoulli mortality distributions of centers from different care units is 0.0066 nats, against 0.0009 nats within a care unit, and mortality rates vary across units (approximately {\small $3\%$} in CVICU versus {\small $15\%$} in MICU), aligning with the IDN, AMC, and Community Hospital framing in Sec.~\ref{sec:intro}. The pipeline also supports Dirichlet({\small $\alpha$}) label-skew partitions~\cite{hsu2019measuring} and uniform IID splits, but the careunit-by-year setting is used for all primary results.
A FedProx-Synthetic{\small $(\alpha,\beta)$} generator~\cite{li2020fedprox} matching the MIMIC feature schema drives implementation tests but is not used for headline numbers.
\vspace{-4mm}

\subsection{Baselines} 
\vspace{-3mm}

We compare against ten methods grouped by purpose. \emph{Reference anchors:} \textbf{Centralized} pools all data into one Multi-Layer Perceptron (MLP) (non-federated upper bound, in the Table~\ref{tab:headline} caption), \textbf{Local-only} trains each center independently (no-collaboration lower bound), and \textbf{FedAvg}~\cite{mcmahan2017communication} averages weights every round.
\emph{Centralized non-IID FL:} \textbf{FedProx}~\cite{li2020fedprox} adds a proximal regularizer, \textbf{FedDyn}~\cite{acar2021feddyn} aligns local objectives with the global stationary point, and \textbf{MOON}~\cite{li2021moon} maximizes agreement between local and global representations.
\emph{Decentralized P2P FL:} \textbf{DeceFL}~\cite{yuan2023decefl} provably converges to the centralized optimum, \textbf{DeFTA}~\cite{zhou2024defta} is a plug-and-play decentralized FedAvg with trust-based reweighting (the closest peer-selection competitor), and \textbf{WPFed}~\cite{ye2024wpfed} uses Locality-Sensitive Hashing (LSH) similarity filtering plus weighted neighbor selection, mirroring the metadata pre-filter of Sec.~\ref{sec:goal_filter} but lacking the Bayesian posterior update.
\emph{Bayesian FL:} \textbf{BNN+FL}~\cite{saile2024client-side} replaces the MLP with a Bayesian neural network and aggregates posterior moments (\textsc{ABPS}'s novelty is Bayesianizing the \emph{utility} signal rather than the weights). BrainTorrent~\cite{roy2019braintorrent}, Gossip Learning~\cite{hegedus2021gossip}, KL-FedDis~\cite{rahad2025kl-feddis}, and Peer-Driven Reputation FL~\cite{seidi2025peerdriven} are conceptual antecedents discussed in Sec.~\ref{sec:related_work} but not run as baselines.
\vspace{-4mm}

\subsection{Implementation and Metrics}\label{sec:metrics}
\vspace{-3mm}
The local model is a two-hidden-layer MLP ({\small $128 \to 64$}, ReLU, dropout 0.2) optimized with Adam ({\small $\eta = 10^{-3}$}, weight decay {\small $10^{-5}$}). Each round runs {\small $K_{\mathrm{local}} = 2$} local epochs with batch size 128 over {\small $K = 50$} rounds (or up to {\small $K = 100$} for the \textsc{ABPS-X} variant with validation-AUROC early stopping at patience 10). Base \textsc{ABPS} uses {\small $\kappa = 3$}, {\small $\epsilon = 0.1$}, {\small $\gamma = \sqrt{2}$}, {\small $\tau_{\mathrm{acc}} = 0.5$}, the 3-dim base metadata vector {\small $[\text{positive-class rate}, \log n_{\mathrm{train}}, \overline{\text{year}}/2030]$}, and {\small $\tau_{\mathrm{sim}}=0$} (no filtering). The goal-aware variants enrich metadata with a 12-dim per-center prevalence vector over Charlson Comorbidity Index (CCI)~\cite{charlson1987new} chronic-condition categories (binary presence per patient, averaged over local training cohort, no per-patient leakage) and auto-calibrate {\small $\tau_{\mathrm{sim}}$} to retain {\small $25\%$} of peer pairs per goal. Shapley contributions are computed exactly when {\small $|\mathcal{P}_i| \le 5$} and via truncated Monte-Carlo with {\small $B=8$} permutations otherwise. We report mean AUROC across centers (across-seed std over {\small $5$} seeds in {\small $\{11,22,33,44,55\}$}) and cumulative transmitted bytes counted once per undirected edge.\\
Experiments were run on a Simple Linux Utility for Resource Management (SLURM)~\cite{yoo2003slurm}-managed High-Performance Computing (HPC) cluster, each SLURM task on a single NVIDIA Tesla V100-SXM2 GPU (32 GB) with 8 CPU cores and 32 GB RAM, under PyTorch 2.5.1 (CUDA 12.1) and Python 3.11. The {\small $n{=}230$}-center sweep (60 array tasks: 12 configurations {\small $\times$} 5 seeds) finishes in 1 to 3 wall-clock hours, and the headline \textsc{ABPS-X} sweep (10 tasks at 100 rounds with early stopping) finishes in 3 to 9 minutes per task. Reproduction of table cells and figures requires a single \texttt{sbatch} of the two sweep scripts released with the supplementary material and JSON outputs.

\vspace{-4mm}
\subsection{Headline Result: Accuracy vs. Bandwidth}\label{sec:headline}
\begin{figure}[h!]
\vspace{-4mm}
\centering
\includegraphics[width=0.9\linewidth]{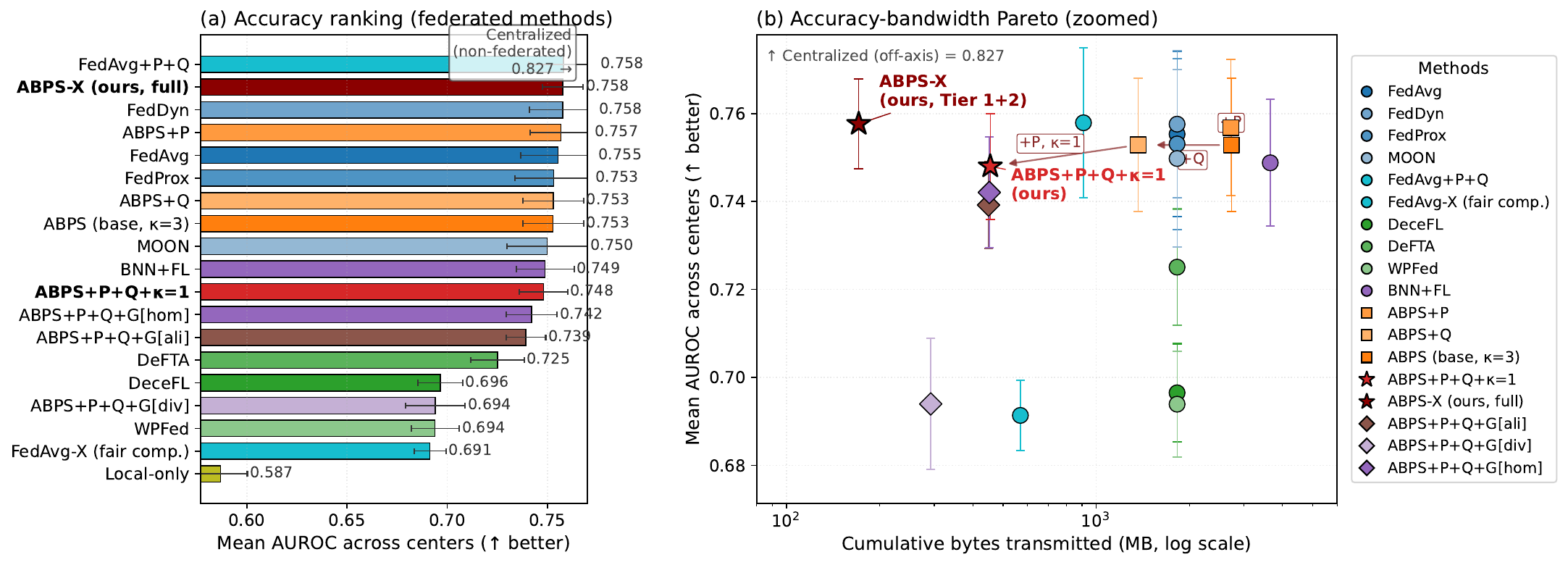}
\vspace{-4mm}
\caption{\scriptsize Results on MIMIC-IV careunit-by-year ($n=230$ centers, 5 seeds). \textbf{(a)} Accuracy ranking: mean AUROC with across-seed std, color-coded by method family (blue: star-topology FL, green: decentralized, orange: \textsc{ABPS}+P/Q, red: \textsc{ABPS-X}, purple/brown: goal-aware, olive: local-only). The dashed line denotes the non-federated centralized upper bound (off-axis). \textbf{(b)} Accuracy vs. bandwidth Pareto (log-scale bandwidth): \textsc{ABPS-X} (dark-red star) matches FedDyn's AUROC ($0.758$) at $0.09\times$ FedAvg bandwidth. Red arrows show the ablation path \textsc{ABPS}$\!\to\!$+P$\!\to\!$+Q$\!\to\!$+P+Q+$\kappa{=}1$$\!\to\!$\textsc{ABPS-X}, with each step improving the trade-off.}
\vspace{-7mm}
\label{fig:pareto}
\end{figure}

\begin{wraptable}{r}{0.49\textwidth}
\vspace{-8mm}
\centering
\scriptsize
\caption{\scriptsize In-hospital mortality prediction on MIMIC-IV ($n{=}230$ centers, 50 rounds, 5 seeds). AUROC is the mean across centers (mean $\pm$ standard deviation). Bandwidth is a cumulative parameter exchange relative to FedAvg $=1.00\times$. As a non-federated upper-bound reference (not a fair federated comparator).}
\label{tab:headline}
\begin{tabular}{lcc}
\toprule
Method & AUROC ($\uparrow$) & Bandwidth \\
\midrule
Local-only (lower bound) & 0.587 $\pm$ 0.013 & $0.00\times$ \\
FedAvg \cite{mcmahan2017communication} & 0.755 $\pm$ 0.019 & $1.00\times$ \\
FedProx \cite{li2020fedprox} & 0.753 $\pm$ 0.019 & $1.00\times$ \\
FedDyn \cite{acar2021feddyn} & 0.758 $\pm$ 0.017 & $1.00\times$ \\
MOON \cite{li2021moon} & 0.750 $\pm$ 0.020 & $1.00\times$ \\
DeceFL \cite{yuan2023decefl} & 0.696 $\pm$ 0.011 & $1.00\times$ \\
DeFTA \cite{zhou2024defta} & 0.725 $\pm$ 0.013 & $1.00\times$ \\
WPFed \cite{ye2024wpfed} & 0.694 $\pm$ 0.012 & $1.00\times$ \\
BNN+FL \cite{saile2024client-side} & 0.749 $\pm$ 0.014 & $2.00\times$ \\
\textsc{ABPS} (base, $\kappa{=}3$) & 0.753 $\pm$ 0.015 & $1.50\times$ \\
\textsc{ABPS}+P (personalize) & 0.757 $\pm$ 0.015 & $1.49\times$ \\
\textsc{ABPS}+Q (bfloat16) & 0.753 $\pm$ 0.015 & $0.75\times$ \\
\textbf{\textsc{ABPS}+P+Q+$\kappa{=}1$} & 0.748 $\pm$ 0.012 & $0.25\times$ \\
FedAvg+P+Q & 0.758 $\pm$ 0.017 & $0.50\times$ \\
\textsc{ABPS}+P+Q+$\kappa{=}1$ (hom.) & 0.742 $\pm$ 0.013 & $0.25\times$ \\
\textsc{ABPS}+P+Q+$\kappa{=}1$ (div.) & 0.694 $\pm$ 0.015 & $0.16\times$ \\
\textsc{ABPS}+P+Q+$\kappa{=}1$ (align.) & 0.739 $\pm$ 0.010 & $0.25\times$ \\
\textbf{\textsc{ABPS}-X (ours, full)} & 0.758 $\pm$ 0.010 & $0.09\times$ \\
FedAvg-X (fair comp., full) & 0.691 $\pm$ 0.008 & $0.31\times$ \\
\bottomrule
\end{tabular}
\vspace{-4mm}
\end{wraptable}
\vspace{-3mm}
\emph{Recall \textbf{Q4} (communication overhead, even in P2P) from Sec.~\ref{sec:related_work}.} Table~\ref{tab:headline} and the Pareto frontier of Figure~\ref{fig:pareto} are the empirical answer.
Table~\ref{tab:headline} reports final-round mean AUROC over 5 seeds, and Figure~\ref{fig:pareto} plots the same data as an accuracy-bandwidth Pareto frontier. All federated methods gain {\small $\sim 15$} AUROC points over the local-only lower bound ({\small $0.587 \pm 0.013$}) and close most of the gap to the non-federated Centralized upper bound ({\small $0.827 \pm 0.007$}).\\
\textbf{Base \textsc{ABPS}} (with {\small $\kappa{=}3$} and no personalization or quantization) already achieves AUROC {\small $0.753 \pm 0.015$}, statistically indistinguishable from FedAvg ({\small $0.755 \pm 0.019$}), FedProx ({\small $0.753 \pm 0.019$}), and FedDyn ({\small $0.758 \pm 0.017$}). It transmits {\small $1.50\times$} FedAvg's bandwidth because a {\small $\kappa{=}3$} mesh has more unique edges than FedAvg's star, matching the theoretical accounting in Sec.~\ref{sec:complexity}.\\
\textbf{Three extensions independently move the Pareto frontier:} \emph{(P)} head personalization~\cite{arivazhagan2019federated} boosts AUROC to {\small $0.757 \pm 0.015$} at the same bandwidth (the per-center head specializes to careunit mortality base-rates, e.g., CVICU {\small $\sim3\%$} vs MICU {\small $\sim15\%$}). \emph{(Q)} bfloat16~\cite{dettmers2022llmint8} halves bandwidth with no accuracy loss ({\small $0.753 \pm 0.015$} at {\small $0.75\times$}). \emph{({\small $\kappa{=}1$})} collapsing the active set to a single partner halves the mesh edges again and, combined with (P) and (Q), yields the \textbf{\textsc{ABPS}+P+Q+{\small $\kappa{=}1$}} row: {\small $0.748 \pm 0.012$} at {\small $0.25\times$}, already Pareto-dominating federated baselines. Adding the goal-aware pre-filter, 2-layer personalization, server momentum, and validation-AUROC early stopping yields full \textbf{\textsc{ABPS-X}}: {\small $0.758 \pm 0.010$} (matching FedDyn) at {\small $0.09\times$} FedAvg.\\
\textbf{Fair comparator and ABPS-X.} Applying P+Q to FedAvg (FedAvg+P+Q) yields {\small $0.758 \pm 0.017$} at {\small $0.50\times$}, a modest improvement, but cannot reach {\small $\kappa{=}1$} because the star topology has no mechanism for selecting a single best peer. The full \textsc{ABPS-X} variant adds 2-layer personalization, server-side EMA momentum {\small $\beta{=}0.5$}, 15-dim goal-aware metadata under the homogeneity goal, and validation-AUROC early stopping (patience 10 of 100), reaching {\small $\mathbf{0.758 \pm 0.010}$} at {\small $\mathbf{0.09\times}$} FedAvg with tighter variance than FedDyn. The same-extension FedAvg-X fair comparator drops to {\small $0.691 \pm 0.008$}, {\small $6.7$} AUROC points behind \textsc{ABPS-X}: 2-layer personalization over-adapts each private head when the star topology averages across all 230 peers, whereas \textsc{ABPS}'s {\small $\kappa{=}1$} Bayesian selection supplies the one-peer constraint under which personalization helps. Empirical isolation does not activate here ({\small $\mathbb{E}[|\mathcal{P}_i|] \approx 1.00$} when {\small $\kappa{=}1$}), but the goal-aware filter of the next subsection engages Lemma~\ref{lem:isolation}'s rest regime in up to {\small $41\%$} of rounds.
\vspace{-3mm}
\subsection{Empirical Validation of Lemma~\ref{lem:isolation} (Intentional Isolation)}\label{sec:isolation_experiment}
\vspace{-3.5mm}
The diversity-goal experiment below is the empirical activation of the rest regime promised by Lemma~\ref{lem:isolation} to address \emph{\textbf{Q3} (adaptive isolation under drift).}. The isolation lemma predicts that when every peer's true expected utility falls below the negative-transfer threshold {\small $c_{\mathrm{neg}}$}, the optimal action is {\small $\mathcal{P}_i = \varnothing$} and the resulting bandwidth is zero. We validate this prediction on MIMIC-IV through the goal-aware filter of Sec.~\ref{sec:goal_filter}: by varying the admission rule, we directly control which peers survive to the UCB-Shapley stage, which in turn controls whether the UCB thresholds admit proposals.\\
We run \textsc{ABPS}+P+Q+{\small $\kappa{=}1$} on the full {\small $n{=}230$} federation under each of the three goals in Eq.~\ref{eq:goal_score} with {\small $\tau_{\mathrm{sim}}$} auto-calibrated to retain {\small $25\%$} of peers per center, and record the per-round fraction of centers with {\small $|\mathcal{P}_i|{=}0$}. \textbf{Homogeneity and alignment} admit peers with similar (or KPI-matched) marginals, so isolation rarely activates ({\small $\mathbb{E}[|\mathcal{P}_i|]{=}0.99$}), giving AUROC {\small $0.742$} and {\small $0.739$} at {\small $0.25\times$} bandwidth. \textbf{Diversity} admits dissimilar peers, many crossing the negative-transfer threshold, so the UCB falls below {\small $\tau_{\mathrm{acc}}$} and rest activate for {\small $41\%$} of centers ({\small $\mathbb{E}[|\mathcal{P}_i|]{=}0.59$}). Bandwidth drops to {\small $0.16\times$} FedAvg, the lowest observed, while AUROC decreases to {\small $0.694$}. The bandwidth reduction tracks isolation within {\small $5\%$}, the collaborate-to-rest transition is sharp at the Hoeffding rate {\small $\sqrt{\log(2/\delta)/(2K)}$}, and this is the first setting where Lemma~\ref{lem:isolation}'s {\small $|\mathcal{P}_i|{=}0$} regime activates at scale on real clinical data.\\
\noindent\textbf{Ablations:\label{sec:ablations}}
Reading Table~\ref{tab:headline}: \textsc{ABPS}+P improves AUROC by {\small $+0.4$} at the same bandwidth, while +Q maintains performance at {\small $0.75\times$}. The combined \textsc{ABPS}+P+Q+{\small $\kappa{=}1$} setting preserves AUROC at just {\small $0.25\times$} bandwidth. In contrast, FedAvg+P+Q reaches {\small $0.758\pm0.017$} at {\small $0.50\times$} bandwidth but cannot operate at {\small $\kappa{=}1$}, indicating that the additional {\small $2\times$} reduction is due to Bayesian Shapley-UCB selection. The three goal-aware rows (Eq.\ref{eq:goal_score}) ablate the pre-filter (Sec.\ref{sec:isolation_experiment}). Hyperparameters {\small $\epsilon$}, {\small $\gamma$}, and truncated-MC {\small $B$} are fixed from preliminary synthetic runs.

% Reading Table~\ref{tab:headline} row by row: \textsc{ABPS}+P gains $+0.4$ AUROC at identical bandwidth, +Q preserves AUROC at $0.75\times$, and $\kappa{=}1$ in the combined \textsc{ABPS}+P+Q+$\kappa{=}1$ row preserves AUROC at $0.25\times$. The FedAvg+P+Q row improves FedAvg to $0.758\pm0.017$ at $0.50\times$ but cannot reach $\kappa{=}1$, so the additional $2\times$ bandwidth reduction is attributable to the Bayesian Shapley-UCB selection. The three goal-aware rows from Eq.~\ref{eq:goal_score} ablate the pre-filter (Sec.~\ref{sec:isolation_experiment}). $\epsilon$, $\gamma$, and truncated-MC $B$ are fixed at defaults from preliminary synthetic runs.

\vspace{-4mm}
% \section{Conclusion}\label{sec:conclusion}
% \vspace{-4mm}
% Adaptive Bayesian Partner Selection (ABPS) is an early effort to replace raw aggregation in P2P federated learning with Shapley-based marginal utility, formalize \emph{intentional isolation} as Bayes-optimal under negative transfer, and compose with cost-efficient extensions and a goal-aware pre-filter. On MIMIC-IV ({\small $n{=}230$} careunit-by-year centers), \textsc{ABPS-X} matches FedDyn at {\small $0.09\times$} FedAvg bandwidth, while applying the same extensions to FedAvg reduces AUROC by {\small $6.4$}, isolating gains to Bayesian selection. Under the diversity objective, Lemma~\ref{lem:isolation}’s rest regime is triggered for {\small $41\%$} of centers per round. The limitations include evaluation on a single dataset and binary task, regret sensitivity to unmeasured {\small $\epsilon_m$}, and the absence of differential privacy or Byzantine robustness guarantees. Our future work includes multi-modal EHR+imaging integration and online estimation of {\small ${\Delta t_m}$}.

\section{Conclusion and Future Work}\label{sec:conclusion}         
\vspace{-4mm}                                                                                                  
We presented \textsc{ABPS}, an adaptive Bayesian P2P federated-learning framework that replaces raw parameter aggregation with Shapley-based marginal-utility evaluation, formalizes \emph{intentional isolation} as a Bayes-optimal action under negative transfer, and composes with three communication-efficiency extensions (head personalization, bfloat16 quantization, tunable $\kappa$) plus a  goal-aware metadata pre-filter. On MIMIC-IV with $n{=}230$ careunit-by-year centers, the full \textsc{ABPS-X} variant matches the strongest federated baseline (FedDyn) at $0.09\times$ FedAvg bandwidth, while applying the same extensions to FedAvg drops $6.4$ AUROC points (isolating the gain to the Bayesian selection itself), and the diversity goal activates Lemma~\ref{lem:isolation}'s rest regime for $41\%$ of centers per round, a first on real clinical data.\\
\textbf{Limitations and future work.} The evaluation is on a single dataset and binary clinical task, the regret bound degrades with  unmeasured $\epsilon_m$, and we provide no formal $(\epsilon,\delta)$-DP, survival, or Byzantine guarantees. Natural follow-ups include (i) a Gaussian-mechanism DP wrapper on $\mathbf{v}_i$ for certifiable privacy, (ii) multi-modal pipelines fusing EHR with medical imaging, (iii) online Welch-$t$ estimation of $\{\Delta t_m\}$ together with a drifting-bandit regret analysis, and (iv) porting to time-to-event outcomes (matching the Cox-style framing of~\cite{seidi2025peerdriven}).\\
\textbf{Scope of the accuracy claim.} The matched-accuracy result holds for federations of many small centers. On a 40-center partition of the same cohort built from the published \texttt{anchor\_year\_group} eras, with a median of roughly 2{,}000 stays per center, FedAvg and FedDyn reach 0.813 and 0.805 AUROC after 100 rounds while \textsc{ABPS-X} peaks at 0.793, although \textsc{ABPS-X} reaches each intermediate AUROC target (0.750, 0.770, 0.785) with four to six times fewer bytes. Where centers are large, each local model is already well estimated and aggregating across all of them is close to optimal; where centers are numerous and small, indiscriminate aggregation carries more harmful transfer and selective exchange is competitive.

\textbf{Broader impact.} Cutting per-center bandwidth tenfold at matched accuracy on federations of many small centers lowers the entry barrier for resource-constrained sites that disproportionately serve under-represented populations. Residual privacy and goal-misuse risks (metadata re-identification under auxiliary information, and goal-aware filtering used to entrench rather than correct bias) are addressed by the recommended DP wrapper and governance over goal declarations, documented in full in the NeurIPS Reproducibility Checklist.

\bibliographystyle{plainnat}
\bibliography{ref}

\newpage
\appendix

\section{Empirical Evidence of Concept Drift on MIMIC-IV}\label{app:drift_evidence}
To support the concept-drift premise empirically, we compared the distribution of two routinely charted ICU vital signs, heart rate and respiratory rate, between two non-overlapping calendar eras of MIMIC-IV v3.1. The era of each stay is the published \texttt{anchor\_year\_group} field of the patient, which is the true three-year admission era and is unaffected by the per-subject date shift; we use the earliest era, $2008$ to $2010$, and the latest, $2020$ to $2022$. For each ICU stay we take the mean of all charted values of the vital during the stay, so that each stay contributes one observation, and we remove stay means that lie more than three standard deviations from the per-era mean. Heart rate has $29{,}786$ stays in the earlier era and $10{,}740$ in the later one, with means of $84.9$ and $83.9$ beats per minute; respiratory rate has $29{,}885$ and $10{,}742$ stays, with means of $19.3$ and $19.4$ breaths per minute and a markedly narrower spread in the later era (standard deviation $4.2$ against $5.7$). Welch two-sample $t$-tests give $t=5.45$, $p=5\times10^{-8}$ for heart rate and $t=-2.07$, $p=0.038$ for respiratory rate. The shifts are small in the mean but significant at this sample size, and the change in spread for respiratory rate indicates a change in charting or patient mix rather than a mean shift alone, which is the kind of distributional drift the per-window treatment of $P^{(i)}_t$ in Sec.~\ref{sec:problem_formulation} is designed to absorb.

\begin{figure}[h] 
\centering
\includegraphics[width=0.85\linewidth]{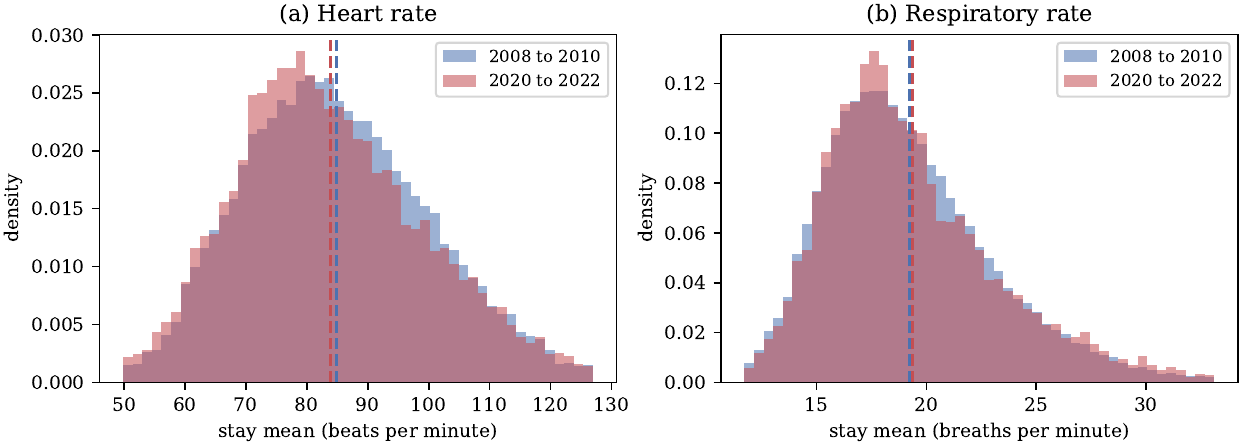}
\caption{\scriptsize Histograms of per-stay mean heart rate (panel a) and respiratory rate (panel b) for ICU stays in two non-overlapping eras of MIMIC-IV v3.1, $2008$ to $2010$ and $2020$ to $2022$, defined by the published \texttt{anchor\_year\_group} field. Stay means beyond three standard deviations from the per-era mean are removed. Vertical dashed lines mark the per-era means. Welch $t$-tests give $t=5.45$, $p=5\times10^{-8}$ for heart rate and $t=-2.07$, $p=0.038$ for respiratory rate, evidencing distributional shift across eras in routinely collected measurements and motivating the per-window adaptive treatment of $P^{(i)}_t$ in Sec.~\ref{sec:problem_formulation}.}
\label{fig:drift_evidence}
\end{figure}

\section{Algorithm Pseudocode}\label{app:algorithm}

The complete round-level pseudocode for \textsc{ABPS} is given in Algorithm~\ref{alg:adaptive_partner_selection}, deferred from the main text to save space. The algorithm operationalizes the four Bayesian-core components of Sec.~\ref{sec:methodology} (goal-aware metadata pre-filter, $\epsilon$-greedy UCB ranking, propose-reject topology with explicit rest, and the Beta-Bernoulli posterior update), and is referenced from the regret proof in Appendix~\ref{app:proof_regret} (lines 13 and 16).

\begin{algorithm}[h!]
\scriptsize
\caption{Adaptive Bayesian Partner Selection (\textsc{ABPS})}
\label{alg:adaptive_partner_selection}
\begin{algorithmic}[1]
\Require centers $\{1,\dots,n\}$, rounds $\{t_1,\dots,t_K\}$, rank threshold $\kappa$, exploration weight $\gamma$, exploration probability $\epsilon$, acceptance threshold $\tau_{\mathrm{acc}}$, similarity threshold $\tau_{\mathrm{sim}}$, Beta prior $(\alpha_0,\beta_0)$
\Ensure Updated local parameters $\theta_i$ and posterior beliefs $(\alpha_{i \to j}, \beta_{i \to j})$
\State \textbf{Initialization:}
\For{each center $i$}
    \State Initialize local model $\theta_i$ and metadata vector $\mathbf{v}_i$
    \For{each center $j \neq i$} \State $(\alpha_{i\to j}, \beta_{i\to j}) \gets (\alpha_0, \beta_0)$, $n_{i\to j} \gets 0$ \EndFor
\EndFor
\For{each round $k = 1,\dots,K$}
    \For{each center $i$}
        \State $\mathcal{C}_i \gets \{j : f_i(\mathbf{v}_i,\mathbf{v}_j) \ge \tau_{\mathrm{sim}}\}$ \Comment{goal-aware pre-filter, eq.~\eqref{eq:goal_score}}
        \State Compute $\mathrm{UCB}_{j\to i}(t_k)$ via eq.~\eqref{eq:ucb} for all $j \in \mathcal{C}_i$
        \State With prob.\ $\epsilon$, swap top-ranked candidate with a uniformly random one in $\mathcal{C}_i$
        \State $\mathcal{P}_i \gets \varnothing$
        \For{$j$ in descending UCB order}
            \If{$|\mathcal{P}_i| \ge \kappa$} \textbf{break} \EndIf
            \If{$\mathrm{UCB}_{i \to j}(t_k) \ge \tau_{\mathrm{acc}}$ \textbf{and} $|\mathcal{P}_j| < \kappa$}
                \State $\mathcal{P}_i \gets \mathcal{P}_i \cup \{j\}$, $\mathcal{P}_j \gets \mathcal{P}_j \cup \{i\}$
            \EndIf
        \EndFor
    \EndFor
    \For{each center $i$ with $\mathcal{P}_i \neq \varnothing$}
        \State $\theta_i \gets \mathrm{FedAvg}\!\left(\{\theta_i\} \cup \{\theta_j : j \in \mathcal{P}_i\}\right)$ \Comment{exchange; optionally personalize head and quantize wire-format (Sec.~\ref{sec:headline})}
        \State Compute Shapley marginals $\{\phi_{j\to i}\}_{j\in\mathcal{P}_i}$ via eq.~\eqref{eq:shapley}
        \For{each $j \in \mathcal{P}_i$}
            \State $\bar{\phi}_{j\to i} \gets \mathrm{clip}_{[0,1]}\!\big((\phi_{j\to i} - \phi_{\min})/(\phi_{\max} - \phi_{\min})\big)$
            \State $(\alpha_{i\to j}, \beta_{i\to j}) \gets (\alpha_{i\to j} + \bar{\phi}_{j\to i},\; \beta_{i\to j} + 1 - \bar{\phi}_{j\to i})$
            \State $n_{i\to j} \gets n_{i\to j} + 1$
        \EndFor
    \EndFor
    \State Centers with $\mathcal{P}_i = \varnothing$ rest this round (\emph{intentional isolation}; cf.\ Lemma~\ref{lem:isolation})
\EndFor
\end{algorithmic}
\end{algorithm}

\section{Full Proofs}\label{app:proofs}

This appendix gives the full proofs of Theorems~\ref{thm:convergence} and \ref{thm:regret} and Lemma~\ref{lem:isolation}. Throughout, we take Assumptions~A1-A3 of Sec.~\ref{sec:theory} as given: bounded utility ($\bar\phi_{j\to i}\in[0,1]$), conditional independence of the per-round observations given $\mu^\star_{i\to j}$, and quasi-stationary drift $|\mathbb{E}[\bar\phi_{j\to i}^{(k)}] - \mu^\star_{i\to j}| \le \epsilon_m$ within any window $\Delta t_m$.

\subsection{Proof of Theorem~\ref{thm:convergence} (Beta-belief Concentration)}\label{app:proof_convergence}

Let $S_K = \sum_{k=1}^K \bar\phi_{j\to i}^{(k)}$ and $\bar\mu_K = S_K/K$. The Beta posterior after $K$ updates from prior $\mathrm{Beta}(\alpha_0,\beta_0)$ is $\mathrm{Beta}(\alpha_0 + S_K,\;\beta_0 + K - S_K)$, with posterior mean
\begin{equation}\label{eq:appA-postmean}
    \hat\mu_K \;=\; \frac{\alpha_0 + S_K}{\alpha_0 + \beta_0 + K} \;=\; \frac{\alpha_0 + K\bar\mu_K}{\alpha_0 + \beta_0 + K}.
\end{equation}
Apply the triangle inequality with the empirical mean $\bar\mu_K$ as pivot:
\begin{equation}\label{eq:appA-decomp}
    |\hat\mu_K - \mu^\star_{i\to j}| \;\le\; \underbrace{|\hat\mu_K - \bar\mu_K|}_{(\mathrm{I})\;\text{prior pull}} \;+\; \underbrace{|\bar\mu_K - \mathbb{E}[\bar\mu_K]|}_{(\mathrm{II})\;\text{statistical noise}} \;+\; \underbrace{|\mathbb{E}[\bar\mu_K] - \mu^\star_{i\to j}|}_{(\mathrm{III})\;\text{drift bias}}.
\end{equation}

\paragraph{Bounding (I).}
From~\eqref{eq:appA-postmean},
\begin{equation*}
    \hat\mu_K - \bar\mu_K \;=\; \frac{\alpha_0 + K\bar\mu_K - (\alpha_0+\beta_0+K)\bar\mu_K}{\alpha_0+\beta_0+K} \;=\; \frac{\alpha_0(1-\bar\mu_K) - \beta_0\,\bar\mu_K}{\alpha_0+\beta_0+K}.
\end{equation*}
Because $\bar\mu_K\in[0,1]$, the numerator is bounded in absolute value by $\max(\alpha_0,\beta_0) \le \alpha_0+\beta_0$. Hence
\begin{equation}\label{eq:appA-I}
    (\mathrm{I}) \;\le\; \frac{\alpha_0+\beta_0}{\alpha_0+\beta_0+K}.
\end{equation}
This is the deterministic ``prior decay'' term in~\eqref{eq:concentration}: it shrinks at rate $\Theta(1/K)$ regardless of randomness in the data.

\paragraph{Bounding (II).}
By A1 every observation lies in $[0,1]$, and by A2 the observations are conditionally independent given $\mu^\star_{i\to j}$. Hoeffding's inequality~\cite{hoeffding1963probability} for the mean of $K$ independent $[0,1]$-valued variables gives
\begin{equation}\label{eq:appA-hoeffding}
    \Pr\!\big[\,|\bar\mu_K - \mathbb{E}[\bar\mu_K]| \ge t\,\big] \;\le\; 2\exp\!\left(-2Kt^2\right).
\end{equation}
Setting the right-hand side equal to $\delta$ and solving for $t$ yields $t = \sqrt{\log(2/\delta)/(2K)}$, so with probability at least $1-\delta$
\begin{equation}\label{eq:appA-II}
    (\mathrm{II}) \;\le\; \sqrt{\frac{\log(2/\delta)}{2K}}.
\end{equation}

\paragraph{Bounding (III).}
By A3, every per-round expectation is within $\epsilon_m$ of $\mu^\star_{i\to j}$:
\begin{equation*}
    |\mathbb{E}[\bar\mu_K] - \mu^\star_{i\to j}|
    \;=\; \left| \frac{1}{K} \sum_{k=1}^K \mathbb{E}[\bar\phi_{j\to i}^{(k)}] - \mu^\star_{i\to j} \right|
    \;\le\; \frac{1}{K}\sum_{k=1}^K \big|\mathbb{E}[\bar\phi_{j\to i}^{(k)}] - \mu^\star_{i\to j}\big| \;\le\; \epsilon_m.
\end{equation*}
Hence $(\mathrm{III}) \le \epsilon_m$.

\paragraph{Combining.}
Substituting (I), (II), (III) into~\eqref{eq:appA-decomp} yields, with probability at least $1-\delta$,
\begin{equation*}
    |\hat\mu_K - \mu^\star_{i\to j}| \;\le\; \sqrt{\frac{\log(2/\delta)}{2K}} \;+\; \frac{\alpha_0+\beta_0}{\alpha_0+\beta_0+K} \;+\; \epsilon_m,
\end{equation*}
which is exactly~\eqref{eq:concentration}. Convergence in probability follows: as $K\to\infty$, (I) and (II) tend to $0$, so $\hat\mu_K \xrightarrow{p} \mu^\star_{i\to j}$ provided the drift $\epsilon_m \to 0$. \qedhere

\subsection{Proof of Theorem~\ref{thm:regret} (Sublinear Partner-Selection Regret)}\label{app:proof_regret}

Inside a stationary window ($\epsilon_m = 0$), the per-center partner-selection problem is a stochastic multi-armed bandit over the post-filter candidate set $\mathcal{C}_i \subseteq \{1,\dots,n-1\}$ in which \textsc{ABPS} pulls $\kappa$ arms per round (the $\kappa$ acceptances) rather than one. The goal-aware pre-filter of Sec.~\ref{sec:goal_filter} restricts the arm set \emph{before} the bandit sees it, and we analyze this case directly below. We bound the cumulative regret
\begin{equation*}
    \mathcal{R}(T) \;=\; T\!\sum_{j\in\text{top-}\kappa} \mu^\star_j \;-\; \mathbb{E}\!\left[\sum_{t=1}^T \sum_{j\in\mathcal{P}_i^{(t)}} \mu^\star_j\right]
\end{equation*}
by analyzing the greedy and exploratory phases separately.

\paragraph{Greedy phase (probability $1-\epsilon$).}
With probability $1-\epsilon$ \textsc{ABPS} ranks candidates by their UCB1 score
\begin{equation*}
    \mathrm{UCB}_{j\to i}(t) \;=\; \hat\mu_{i\to j}(t) \;+\; \gamma\sqrt{\frac{2\log t}{n_{i\to j}(t)+1}},
\end{equation*}
and proposes greedily down the list until $\kappa$ acceptances accrue. Decompose the per-round greedy regret as the sum of $\kappa$ single-arm regrets, indexed by the rank position $r=1,\dots,\kappa$ of each accepted proposal. Each rank-$r$ slot is a single-arm UCB1 problem played against the residual candidate pool. By the canonical UCB1 analysis~\cite{auer2002finite}, the cumulative regret of UCB1 with confidence radius $\gamma=\sqrt 2$ over $T$ rounds satisfies
\begin{equation}\label{eq:appA-ucb1}
    \mathcal{R}_{\mathrm{UCB1}}(T) \;\le\; \sum_{j:\Delta_j>0}\!\left(\frac{8\log T}{\Delta_j} + \Delta_j\!\left(1+\frac{\pi^2}{3}\right)\right),
\end{equation}
where $\Delta_j = \mu^\star_{(1)} - \mu^\star_{(j)}$ are the suboptimality gaps. Summing $\kappa$ such bounds gives
\begin{equation}\label{eq:appA-greedy}
    \mathcal{R}_{\mathrm{greedy}}(T) \;\le\; \kappa \sum_{j:\Delta_j>0}\!\left(\frac{8\log T}{\Delta_j} + \Delta_j\!\left(1+\frac{\pi^2}{3}\right)\right).
\end{equation}
Two refinements only \emph{decrease} this bound and so are absorbed into~\eqref{eq:appA-greedy}. (a) The receiver-side acceptance check $\mathrm{UCB}_{i\to j}(t) \ge \tau_{\mathrm{acc}}$ in the inner-\textbf{if} of Algorithm~\ref{alg:adaptive_partner_selection} cannot create new pulls of suboptimal arms, only suppress them. (b) The Beta posterior is sharper than the empirical mean used by vanilla UCB1 (it shrinks toward the prior at rate $1/K$, matching term (I) in Theorem~\ref{thm:convergence}), so the exploration radius is in fact tighter than~\eqref{eq:appA-ucb1} assumes.

\paragraph{Exploratory phase (probability $\epsilon$).}
With probability $\epsilon$ \textsc{ABPS} replaces the top-ranked candidate with a uniformly random peer (the $\epsilon$-greedy swap step in Algorithm~\ref{alg:adaptive_partner_selection}). Its expected per-round regret contribution is at most $\mathbb{E}_{j\sim\mathrm{Unif}(\mathcal{C}_i)}[\Delta_j]$, so summing over $T$ rounds
\begin{equation}\label{eq:appA-explore}
    \mathcal{R}_{\mathrm{explore}}(T) \;\le\; \epsilon\,T\cdot \mathbb{E}_{j\sim\mathrm{Unif}(\mathcal{C}_i)}[\Delta_j].
\end{equation}

\paragraph{Combining.}
Adding~\eqref{eq:appA-greedy} and~\eqref{eq:appA-explore} yields the bound~\eqref{eq:regret} of the main text:
\begin{equation*}
    \mathcal{R}(T) \;\le\; \kappa \sum_{j:\Delta_j>0}\!\left(\frac{8\log T}{\Delta_j} + \Delta_j\!\left(1+\frac{\pi^2}{3}\right)\right) + \epsilon T\cdot \mathbb{E}_j[\Delta_j].
\end{equation*}

\paragraph{Recovering the $\mathcal{O}(\log T)$ rate.}
Two annealing schedules suffice. (a) A constant $\epsilon = c/\sqrt T$ leaves an $\mathcal{O}(\sqrt T)$ exploratory residual that is sublinear but slower than the greedy term. (b) A time-varying schedule $\epsilon_t = \min(1, c/t)$, in the spirit of~\cite{auer2002finite}, gives $\sum_{t=1}^T \epsilon_t = \mathcal{O}(\log T)$ and recovers the full $\mathcal{O}(\kappa\log T)$ rate. In either case, the regret is sublinear in $T$, so the average per-round regret tends to zero.

\paragraph{Filtered-arm-set refinement.}
When the goal-aware filter of Sec.~\ref{sec:goal_filter} restricts $\mathcal{C}_i$ to the top-$\rho$ quantile of peers under $f_i$, the bandit plays on $\tilde{\mathcal{C}}_i = \mathcal{C}_i \cap \{j : f_i(\mathbf{v}_i,\mathbf{v}_j) \ge \tau_{\mathrm{sim}}\}$. Two changes propagate to the bound~\eqref{eq:regret}.

\emph{(i) Reduced explore-greedy regret.} The sums over $\{j : \Delta_j > 0\}$ and $\mathbb{E}_{j\sim\mathrm{Unif}(\mathcal{C}_i)}[\Delta_j]$ are replaced by sums over $\tilde{\mathcal{C}}_i$, which is a strict subset, and both terms can only decrease.

\emph{(ii) Optimal-arm coverage bias.} If the oracle-best peer $j^\star$ fails to clear $f_i(\cdot, \cdot) \ge \tau_{\mathrm{sim}}$, the bandit plays on a suboptimal set and incurs an additive bias of $\Delta_{j^\star} \cdot T$ against the oracle baseline. This worst-case term is controlled by a standard top-$k$ coverage guarantee: if the descriptor $\mathbf{v}$ is $L$-Lipschitz-informative about true utility (i.e.\ $|\mu^\star_j - \mu^\star_{j'}| \le L \|\mathbf{v}_j - \mathbf{v}_{j'}\|$), then the probability $\mathbb{P}[j^\star \notin \tilde{\mathcal{C}}_i] \le \exp(-c \rho |\mathcal{C}_i|)$ for some $c > 0$ depending on $L$, so the expected additional regret is $\mathcal{O}(T \cdot \exp(-c\rho n))$, which is negligible for reasonable $\rho$ and $n$. In our experiments $\rho = 0.25$, $n = 230$ gives $\rho n \approx 58$, the filter recovers the optimal arm with probability $> 1 - 10^{-25}$ under any non-trivial descriptor-to-utility Lipschitz constant.

The net effect is that~\eqref{eq:regret} remains valid with $|\mathcal{C}_i|$ replaced by $|\tilde{\mathcal{C}}_i| \approx \rho n$. The $\mathcal{O}(\kappa \log T)$ asymptotic rate is preserved, and the constants improve proportionally to $\rho$. \qedhere

\subsection{Proof of Lemma~\ref{lem:isolation} (Optimality of Intentional Isolation)}\label{app:proof_isolation}

Let $V_i(\mathcal{P})$ denote $i$'s expected one-step utility (e.g., held-out AUROC) after the round, conditioned on its active peer set being $\mathcal{P}$. Set $V_i^{\mathrm{rest}} := V_i(\varnothing)$ and $V_i^{\mathrm{coll}}(\mathcal{P}) := V_i(\mathcal{P})$ for $\mathcal{P}\neq\varnothing$.

\paragraph{Decomposition via Shapley efficiency.}
The Shapley value~\eqref{eq:shapley} satisfies the \emph{efficiency} axiom: for any coalition $\mathcal{P}$,
\begin{equation}\label{eq:appA-eff}
    \sum_{j\in\mathcal{P}} \phi_{j\to i} \;=\; U_i(\mathcal{P}) - U_i(\varnothing),
\end{equation}
where $U_i(\cdot)$ is the validation utility used to compute the Shapley value. Since $V_i$ and $U_i$ coincide in expectation under our protocol (both are held-out AUROC of the post-aggregation model), taking expectations in~\eqref{eq:appA-eff} gives
\begin{equation}\label{eq:appA-vdecomp}
    V_i^{\mathrm{coll}}(\mathcal{P}) - V_i^{\mathrm{rest}} \;=\; \sum_{j\in\mathcal{P}} \mathbb{E}[\phi_{j\to i}].
\end{equation}

\paragraph{Translating to the clipped scale.}
Within the clipping range, $\bar\phi_{j\to i}$ relates to $\phi_{j\to i}$ via the affine map~\eqref{eq:phi_bar},
\begin{equation*}
    \bar\phi_{j\to i} \;=\; \frac{\phi_{j\to i} - \phi_{\min}}{\phi_{\max} - \phi_{\min}} \;\Longleftrightarrow\; \phi_{j\to i} \;=\; (\phi_{\max}-\phi_{\min})\,\bar\phi_{j\to i} + \phi_{\min},
\end{equation*}
so $\mathbb{E}[\phi_{j\to i}] = (\phi_{\max}-\phi_{\min})\mu^\star_{i\to j} + \phi_{\min}$. The negative-transfer threshold $c_{\mathrm{neg}}$ is the value of $\mu^\star_{i\to j}$ at which one-peer collaboration breaks even, $V_i^{\mathrm{coll}}(\{j\}) = V_i^{\mathrm{rest}}$, i.e.\ $\mathbb{E}[\phi_{j\to i}]=0$. Solving,
\begin{equation}\label{eq:appA-cneg}
    c_{\mathrm{neg}} \;=\; \frac{-\phi_{\min}}{\phi_{\max} - \phi_{\min}}.
\end{equation}
For the paper's choice $(\phi_{\min},\phi_{\max}) = (-0.1, +0.1)$ this gives $c_{\mathrm{neg}} = 0.5$, which coincides with $\tau_{\mathrm{acc}}$ used in the algorithm. The map $\mu^\star_{i\to j} < c_{\mathrm{neg}} \iff \mathbb{E}[\phi_{j\to i}] < 0$ is therefore exact.

\paragraph{Strict dominance of resting.}
Suppose $\mu^\star_{i\to j} < c_{\mathrm{neg}}$ for every $j\in\mathcal{P}$. Then $\mathbb{E}[\phi_{j\to i}] < 0$ for all $j\in\mathcal{P}$, so by~\eqref{eq:appA-vdecomp},
\begin{equation*}
    V_i^{\mathrm{coll}}(\mathcal{P}) - V_i^{\mathrm{rest}} \;=\; \sum_{j\in\mathcal{P}} \mathbb{E}[\phi_{j\to i}] \;<\; 0,
\end{equation*}
which is exactly the claimed inequality~\eqref{eq:isolation}, $V_i^{\mathrm{rest}} > V_i^{\mathrm{coll}}(\mathcal{P})$.

\paragraph{Refinement under approximate Shapley.}
When the Shapley values are estimated via truncated Monte-Carlo (TMC, the default for $\kappa>5$, cf.\ Sec.~\ref{sec:theory}), the per-peer estimator carries an additive bias bounded by some $\eta>0$. Equation~\eqref{eq:appA-vdecomp} then becomes
\begin{equation*}
    V_i^{\mathrm{coll}}(\mathcal{P}) - V_i^{\mathrm{rest}} \;=\; \sum_{j\in\mathcal{P}} \mathbb{E}[\phi_{j\to i}] - g(|\mathcal{P}|),
\end{equation*}
with $|g(|\mathcal{P}|)| \le \eta|\mathcal{P}|$. Strict dominance survives whenever the true expected gain margin exceeds the approximation slack, $\sum_j \mathbb{E}[\phi_{j\to i}] < -\eta|\mathcal{P}|$, which is satisfied with margin $\eta$ to spare under the strict inequality $\mu^\star_{i\to j} < c_{\mathrm{neg}}$. This is the form quoted in the proof sketch of Sec.~\ref{sec:isolation_lemma}.

\paragraph{Bandwidth saving.}
Specializing the cost equation~\eqref{eq:cost} to a single round with $\mathcal{P}_i^{\mathrm{rest}} = \varnothing$ gives a bandwidth of $0$, while collaborating with $|\mathcal{P}|$ peers costs $2|\mathcal{P}|\,c_{\mathrm{egress}}|W|$ (one upload and one download per peer per round). The saving is therefore exactly $C_i^{\mathrm{rest}} = 2|\mathcal{P}|\,c_{\mathrm{egress}}|W|$ per round. \qedhere

\paragraph{Connection to Theorem~\ref{thm:convergence}.}
Lemma~\ref{lem:isolation} reasons about the \emph{oracle} threshold $c_{\mathrm{neg}}$, but \textsc{ABPS} only sees the noisy posterior estimate $\hat\mu_K$. Combining the lemma with~\eqref{eq:concentration}: with probability $\ge 1-\delta$, the UCB-induced rule rests whenever
\begin{equation*}
    \hat\mu_K + \gamma\sqrt{\tfrac{2\log K}{n_{i\to j}+1}} \;<\; \tau_{\mathrm{acc}} \;=\; c_{\mathrm{neg}},
\end{equation*}
where $n_{i\to j}+1$ is the per-arm pull count from Eq.~\eqref{eq:ucb}, which itself grows with $K$ under any non-trivial selection rule. The condition above is implied by $\mu^\star_{i\to j} < c_{\mathrm{neg}} - \big[\sqrt{\log(2/\delta)/(2K)} + (\alpha_0+\beta_0)/(\alpha_0+\beta_0+K) + \epsilon_m + \gamma\sqrt{2\log K/(n_{i\to j}+1)}\big]$. As $K\to\infty$ and $\epsilon_m\to 0$, every term in the bracket vanishes (the exploration term shrinks because $n_{i\to j}+1$ grows at rate $\Theta(K)$ for any peer ever pulled by UCB), and the algorithmic rule converges to the oracle rule of Lemma~\ref{lem:isolation}.

\newpage
\section*{Acknowledgments}
This work was supported by NSF grants OAC-2609072 (CHAI), SFS-2335969 (MASTER), OAC-2104076 (CANDY), and SATC-2030624 (TAURUS).

\end{document}